\documentclass{article}

\usepackage[nonatbib,final]{neurips_2022}

\usepackage{caption}
\usepackage[T1]{fontenc}
\usepackage[utf8]{inputenc}
\usepackage{babel}
\usepackage[font=small,labelfont=bf]{caption}

\newcommand{\name}{\texttt{DeRy}}
\usepackage[utf8]{inputenc} 
\usepackage[T1]{fontenc}    
\usepackage{colortbl}
\usepackage{algorithm}
\usepackage{algpseudocode}
\usepackage{wrapfig}
\usepackage{float}
\usepackage{hyperref}       
\usepackage{url}            
\usepackage{booktabs}       
\usepackage{amsfonts}       
\usepackage{nicefrac}       
\usepackage{microtype}      
\usepackage{graphicx}

\newcommand{\DQ}[1]{\textcolor{cyan}{[DQ: #1]}}

\newcommand{\typo}[1]{\textcolor{black}{#1}}

\usepackage{bbding}
\usepackage[dvipsnames]{xcolor}
\usepackage{tikz}
\usepackage{subfig}

\usepackage{pbox}
\usepackage{hyperref}
\usepackage{enumitem}
\usepackage{algorithm} 
\usepackage{algpseudocode} 
\usepackage{multirow}
\usepackage{comment}
\usepackage{amsmath,amssymb} 
\usepackage{color}
\title{Deep Model Reassembly}

\author{%
   Xingyi Yang$^1$ \quad Daquan Zhou$^{1,2}$  \quad Songhua Liu$^1$ \quad Jingwen Ye$^1$ \quad Xinchao Wang$^1$\\
   $^1$National University of Singapore \quad $^2$Bytedance \\
   \texttt{\{xyang,daquan.zhou,songhua.liu\}@u.nus.edu,
   \{jingweny,xinchao\}@nus.edu.sg}\\
}

\begin{document}

\maketitle

\begin{abstract}

In this paper, we explore a novel knowledge-transfer task, termed as Deep  Model Reassembly~(\name), for general-purpose model reuse.
Given a collection of heterogeneous models pre-trained from distinct sources and with diverse architectures, the goal of~\name, as its name implies, is to first dissect each model into distinctive building blocks, and then selectively reassemble the derived blocks to produce customized networks under both the hardware resource and performance constraints. Such ambitious nature of~{\name} inevitably imposes significant challenges, including, in the first place, the feasibility of its solution. We strive to showcase that, through a dedicated paradigm proposed in this paper, {\name} can be made not only possibly but practically efficient. Specifically, we conduct the partitions of all pre-trained networks jointly via a cover set optimization, and derive  a number of \emph{equivalence set}, within each of which the network blocks are treated as functionally equivalent and hence interchangeable. 
The equivalence sets learned in this way, in turn, enable  picking and assembling blocks to customize networks subject to certain constraints, which is achieved via solving an integer program backed up with a training-free proxy to estimate the task performance. The reassembled models, 
give rise to
gratifying performances with the user-specified constraints satisfied.
We demonstrate that on ImageNet, the best reassemble model achieves $78.6\%$ top-1 accuracy without fine-tuning, which could be further elevated to $83.2\%$ with end-to-end training.
Our code is available at \url{https://github.com/Adamdad/DeRy}.

\end{abstract}

\section{Introduction}

The unprecedented advances of deep learning and its pervasive impact across various domains are partially attributed to, among many other factors,  the numerous \emph{pre-trained models} released online.
Thanks to the generosity of our community, models of diverse architectures specializing in the same or distinct tasks can be readily downloaded and executed in a plug-and-play manner, which, in turn, largely alleviates the model reproducing effort. The sheer number of pre-trained models also enables extensive knowledge transfer tasks, such as knowledge distillation, in which the pre-trained models
can be reused to produce lightweight or multi-task students.

In this paper, we explore a novel knowledge transfer task, which we coin as \emph{Deep Model Reassembly}~(\name). Unlike most prior  tasks that largely focus on reusing pre-trained models as a whole,  {\name}, as the name implies, goes deeper into the building blocks of pre-trained networks. Specifically, given a collection of such pre-trained heterogeneous models or \emph{Model Zoo}, {\name} attempts to first dissect the pre-trained models into building blocks and then reassemble the building blocks to tailor models subject to  users' specifications, like the computational  constraints of the derived network. 
As such,  apart from
the flexibility for model customization, 
{\name} is expected to
aggregate knowledge from heterogeneous models without increasing computation cost,
thereby preserving or even enhancing the downstream performances.

Admittedly, the nature of {\name} \emph{per se} makes it a highly challenging and ambitious task; in fact, it is even unclear whether a solution is feasible, given that no constraints are imposed over the model architectures in the model zoo. Besides, the reassembly process, which assumes the building blocks can be extracted in the first place, calls for a lightweight strategy to approximate the model performances without re-training, since the reassembled model, apart from the parametric constraints, is expected to behave reasonably well.

\begin{figure}[t]
    \centering
    \includegraphics[width=1.0\linewidth]{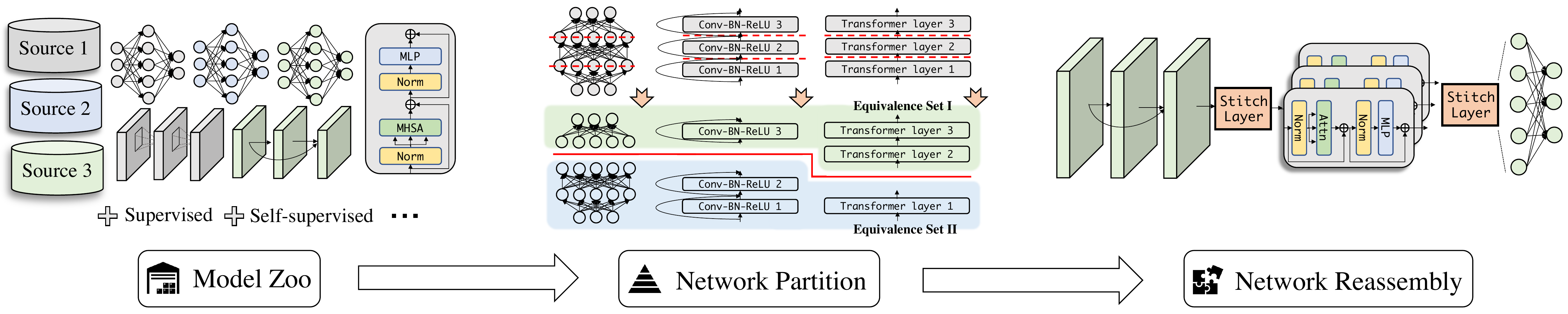}
    \caption{Overall workflow of~\name. It partitions pre-trained models into equivalent sets of neural blocks and then reassemble them for downstream  transfer.     Both steps are optimized through solving constrained programs.}
    \label{fig:DeRy}
    \vspace{-4mm}
\end{figure}

We demonstrate in this paper that, through a dedicated optimization paradigm,  {\name} can be made not only possible by highly efficient. At the heart of our approach is a two-stage strategy that first partitions pre-trained networks into building blocks to form \emph{equivalence sets}, and then selectively assemble building blocks to customize tailored models. Each {equivalence set}, specifically, comprises various building blocks extracted from heterogeneous pre-trained models, which are treated to be functionally equivalent and hence interchangeable. Moreover, the optimization of the two steps is purposely decoupled, so that once the {equivalence sets} are obtained and fixed, they can readily serve as the basis for future network customization.

We show the overall workflow of the proposed {\name} in Figure~\ref{fig:DeRy}. 
It starts by  dissecting pre-trained models into
disjoint sets of neural blocks through solving a cover set optimization problem,
and  derives a number of {equivalence sets}, 
within each of which the neural blocks are treated
as functionally swappable. 
In the second step, {\name} searches for the optimal block-wise reassembly in a training-free manner. 
Specifically, the transfer-ability of a candidate reassembly is estimated by counting the number of linear regions in feature representations~\cite{mellor2021neural}, which reduces the searching cost by $10^4$ times as {compared to training all models exhaustively.}

The reassembled networks, 
apart from satisfying the user-specified hard constraints, 
give rise to truly encouraging results.
We demonstrate through  experiments
that, the reassembled model
achieves $>78\%$ top-1 accuracy on Imagenet
with all blocks frozen.
If we allow for finetuning, the performances can
be further elevated,
sometimes even surpassing 
any pre-trained network in the model zoo.
This phenomenon showcases that  {\name}
is indeed able to aggregate knowledge from various models
and enhance the results.
Besides, {\name} imposes no constraints on the 
network architectures in the model zoo,
and may therefore readily handle various backbones
such as CNN, transformers, and MLP.

Our contributions are thus summarized as follows.
\begin{enumerate}[itemsep=2pt,parsep=2pt,leftmargin=1em,labelwidth=1em,labelsep=2pt]
  \item {We explore a new knowledge transfer task termed Deep Model Reassembly~(\name), 
  which enables reassembling customized networks from a zoo of pre-trained models under
  user-specified constraints.}
  \item {We introduce a novel  two-stage strategy towards solving  {\name},
  by first partitioning the networks into equivalence sets 
  and then reassembling neural blocks to customize networks.
  The two steps are modeled and solved using constrained programming, backed up with
  training-free performance approximations that significantly speed up the knowledge-\typo{transfer} process.
  } 
  \item {The proposed approach achieves competitive performance on a series of transfer learning benckmarks, sometimes even surpassing than any candidate in the model zoo,
  which, in turn, sheds light on the the universal connectivity among pre-trained neural networks.}
\end{enumerate}

\section{Related Work}
\noindent\textbf{Transfer learning from Model Zoo.} 
A standard deep transfer learning paradigm is to leverage a single trained neural network and fine-tune the model on the target task~\cite{xuhong2018explicit,yang2020transfer,li2019delta,howard2018universal,zhou2021deepvit, zhou2020rethinking, hou2021coordinate} or impart the knowledge to other models~\cite{hinton2015distilling, yang2022factorizing, sanh2019distilbert,Yang_2020_CVPR,ye2019student,yang2020factorizable,LiuHuihui21}.
The availability of large-scale model repositories brings about a new problem of transfer learning from a model zoo rather than with a single model. Currently, there are three major solutions. One line of works focuses on select one best model for deployment, either by exhaustive fine-tuning~\cite{kolesnikov2020big,steiner2021train, yang2020transfer} or quantifying the model transferability~\cite{zhang2021quantifying, you2021logme,nguyen2020leep, tran2019transferability,bao2019information,SongJieNuerIPS19,tran2019transferability,bolya2021scalable,kornblith2019better} on the target task. However, due to the unreliable measurement of transferability, the best model selection may be inaccurate, possibly resulting in a suboptimal solution. 
The second idea was to apply ensemble methods~\cite{dietterich2000ensemble,zhou2021ensemble,agostinelli2021transferability, zhou2021autospace}, which inevitably leads to prohibitive computational costs at test time.
The third approach is to adaptively fuse multiple pre-trained models into a single target model. However, those methods can only combine identical~\cite{shu2021zoo,dai2011greedy,wang2020federated} or \typo{homogeneous~\cite{singh2020model,nguyen2021model} network structures}, whereas most model zoo contains diverse architectures. In contrast to standard approaches in Table~\ref{tab:Compare}, \name~dissects the pre-trained models into building blocks and rearranges them in order to reassemble new pre-trained models.
\renewcommand{\arraystretch}{0.9}
\setlength{\tabcolsep}{1pt}
\begin{table}[t]
    \centering
    \footnotesize
    \begin{tabular}{l|c|c|c|c|c}
    \toprule
        Problem & \parbox{1.7cm}{\centering No need to retrain} & \parbox{1.7cm}{\centering Adaptive Architecture} & \parbox{1.9cm}{\centering No Additional Computation} & \parbox{1.8cm}{\centering Utilize All Models} & \parbox{1.9cm}{\centering Heterogeneous Architecture} \\
        \hline
         Single Model Transfer &  \textcolor{Green}{\Checkmark}& \textcolor{red}{\XSolidBrush} & \textcolor{Green}{\Checkmark} & \textcolor{red}{\XSolidBrush} & \textcolor{red}{\XSolidBrush} \\\hline
       Zoo Transfer by Selection &\textcolor{Green}{\Checkmark} & \textcolor{red}{\XSolidBrush} &\textcolor{Green}{\Checkmark} &\textcolor{red}{\XSolidBrush}& \textcolor{Green}{\Checkmark}\\
        Zoo Transfer by Ensemble  & \textcolor{Green}{\Checkmark} & \textcolor{red}{\XSolidBrush} &\textcolor{red}{\XSolidBrush} & \textcolor{Green}{\Checkmark} & \textcolor{Green}{\Checkmark}\\
        Zoo Transfer by Parameter Fusion & \textcolor{Green}{\Checkmark} & \textcolor{red}{\XSolidBrush} &\textcolor{Green}{\Checkmark}& \textcolor{Green}{\Checkmark}& \textcolor{red}{\XSolidBrush}\\
         \hline
        Neural Architecture Search  & \textcolor{red}{\XSolidBrush} &\textcolor{Green}{\Checkmark} & - & - & - \\\hline
       \name & \textcolor{Green}{\Checkmark} & \textcolor{Green}{\Checkmark} & \textcolor{Green}{\Checkmark} & \textcolor{Green}{\Checkmark} & \textcolor{Green}{\Checkmark}\\
       \bottomrule
    \end{tabular}
    \caption{Comparison of a series of transfer learning tasks and our proposed Deep Model Reassembly. } \label{tab:Compare}
    \vspace{-4mm}
\end{table}\\
\noindent\textbf{Neural Representation Similarity.} Measuring similarities between deep neural network representations provide a practical tool to investigate the forward dynamics of deep models. 
Let $\mathbf{X} \in \mathbb{R}^{n\times d_1}$  and $\mathbf{Y} \in \mathbb{R}^{n\times d_2}$ denote two activation matrices for the same $n$ examples. A neural \textit{similarity index} $s(X, Y)$ is a scalar to measure the representations similarity between 
$X$ and $Y$, although they do not necessarily satisfy the triangle inequality required of a proper metric. Several methods including linear regression~\cite{yamins2014performance, hinton2015distilling}, canonical correlation analysis~(CCA)~\cite{ramsay1984matrix,hardoon2004canonical,raghu2017svcca}, centered kernel alignment~(CKA)~\cite{kornblith2019similarity}, generalized
shape metrics~\cite{williams2021generalized}. In this study, we leverage the representations similarity towards function level to quantify the distance between two neural blocks. 
\\
\textbf{Network Stitching.} Initially proposed by~\cite{lenc2015understanding}, model stitching aims to ``plug-in'' the bottom layers of one network into the top layers of another network, thus forming a stitched network~\cite{bansal2021revisiting,csiszarik2021similarity}. It provides an alliterative approach to investigate the representation similarity and invariance of neural networks. A recent line of work achieves competitive performance by stitching a visual transformer on top of the ResNet~\cite{steiner2021train}. Instead of stitching two identical-structured networks in a bottom-top manner, in our study, we investigate to assemble arbitrary pre-trained networks by model stitching.

\section{Deep Model Reassembly}
In this section, we dive into the proposed \name. We first formulate \name, and then define the functional similarity and equivalent sets of neural blocks to partition networks by maximizing overall groupbility. The resulting neural blocks are then linked by solving an integer program.
\subsection{Problem Formulation}
Assume we have a collection of $N$ pre-trained deep neural network models $Z = \{\mathcal{M}_i\}_{i=1}^N$ that each composed of $L_i \in \mathbb{N}$ layers of operation $\{F^{(k)}_i\}_{l=1}^{L_i}$, therefore $\mathcal{M}_i = F^{(1)}_i \circ F^{(2)}_i \dots \circ F^{(L_i)}_i$. Each model can be trained on different tasks or with varied structures. We call $Z$ a \emph{Model Zoo}. We define a learning task $T$ 
composed of a labeled training set $D_{tr} = \{\mathbf{x}_j, y_j\}_{j=1}^M$ and a test set $D_{ts} = \{\mathbf{x}_j\}_{j=1}^L$. 

\textbf{Definition 1} \textit{(Deep Model Reassembly) Given a task $T$, our goal is to find the best-performed $L$-layer compositional model $\mathcal{M}^*$ on $T$, subject
to hard computational or parametric constraints.}

\begin{figure}
\begin{minipage}[c]{.53\textwidth}
\centering
     \includegraphics[width=1.03\linewidth]{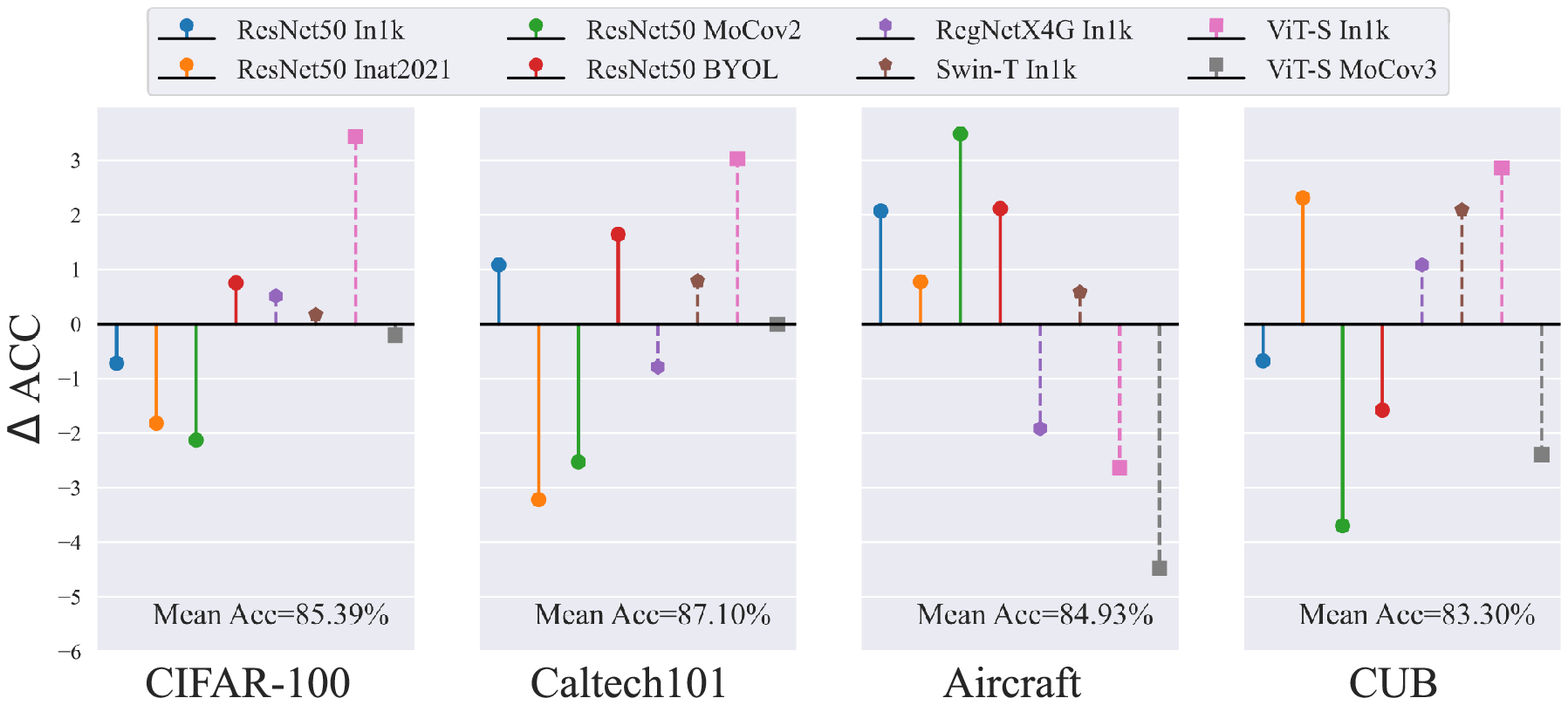}
        \caption{The top-1 accuracy difference between “off-the-shelf” pre-trained models on 4 down-stream tasks.}
        \label{fig:noonewins}
    \end{minipage}%
    \hfill
     \begin{minipage}[c]{.43\textwidth}
        \renewcommand{\arraystretch}{1.1}
        \setlength{\tabcolsep}{0pt}
        \scriptsize
    \begin{tabular}{c|c|c|c}
    \toprule
        Backbone & Init. & \#Params(M) & Acc(\%)  \\
         \midrule
        \multirow{2}{*}{ResNet50} & in1k sup &23.71 & 84.67 \\
         & inat2021 sup & 23.71 & 82.57  \\
        \midrule
        ResNet50 & \pbox{2cm}{\centering \tiny{inat2021(Stage 1\&2) in1k(Stage 3\&4)}} & 23.98 & \textbf{85.30} \\ \midrule\midrule
        ResNet50 & \multirow{2}{*}{in1k sup} &23.71 & 84.67 \\
        Swin-T &  &27.60 & 85.56\\
        \midrule
      \pbox{2cm}{\centering\tiny{ResNet50(Stage 1\&2) 
      Swin-T(Stage 3\&4)}} & in1k sup & 27.94 & \textbf{85.77} \\
        \bottomrule
    \end{tabular}
         \renewcommand\figurename{Table}
    \caption{Accuracy on CIFAR-100 with the pre-trained networks and their reassembled ones. }
    \label{tab:cifar100}
    \end{minipage}%
\end{figure}

We therefore formulate it as an optimization problem
{\footnotesize
\begin{align}
    \mathcal{M}^* = \max_{\mathcal{M}} & P_{T}(\mathcal{M}), \quad s.t. \mathcal{M}=F^{(l_1)}_{i_1} \circ F^{(l_2)}_{i_2} \dots \circ F^{(l_L)}_{i_L}, |\mathcal{M}|\leq C\label{eq:dery}
\end{align}}
where $F^{(l)}_{i}$ is the $l$-th layer of the $i$-th model,  $P_{T}(\mathcal{M})$ indicates the performance on $T$,
and $|\mathcal{M}|\leq C$ denotes the constraints. For two consecutive layers with dimension mismatch, we add a single stitching layer with $1\times 1$ convolution operation to adjust the feature size. The stitching layer structure is described in Supplementary. \\
\textbf{No Single Wins For All.} Figure~\ref{fig:noonewins} provides a preliminary experiment that 8 different pre-trained models are fine-tuned on 4 different image classification tasks. It is clear that no single model universally dominants in transfer evaluations. 
It builds up our primary motivation to reassemble trained models rather than trust the ``best'' candidate.\\
\textbf{Reassembly Might Win.} Table~\ref{tab:cifar100} compares the test performance between the reassembled model and its predecessors. The bottom two stages of the ResNet50 iNaturalist2021~(inat2021 sup)~\cite{van2021benchmarking} are stitched with ResNet50 ImageNet-1k~(in1k sup) stage 3\&4 to form a new model for fine-tuning on CIFAR100. This reassembled model improves its predecessors by 0.63\%/2.73\% accuracy respectively. Similar phenomenon is observed on the reassembled model between ResNet50 in1k and Swin-T in1k. Despite its simplicity, the experiment provides concrete evidence that the neural network reassembly could possibly lead to better model in knowledge transfer.\\
\textbf{Reducing the Complexity.} From the overall $M=\sum_{i=1}^N L_i$ layers, the search space of Eq~\ref{eq:dery} is of size L-permutations of M $P(M, L)$, which is undesirably large. To reduce the overall search cost, we intend to partition the networks into blocks rather than the layer-wise-divided setting. Moreover, it is time-consuming to evaluate each model on the target data through full-time fine-tuning. Therefore, we hope to accelerate the model evaluation, even without model training.

Based on the above discussion, the essence of \name~lies in two steps (1) Partition the networks into blocks and (2) Reassemble the factorized neural blocks. In the following sections, we elaborate on \emph{``what is a good partition?''} and \emph{``what is a good assembly?''}.

\subsection{Network Partition by Functional Equivalence}
\label{sec:funcsim}
A network partition~\cite{feder1999complexity, feder1999complexity} is a division of a neural network into disjoint sub-nets. In this study, we refer specifically to the partition of neural network $\mathcal{M}_i$ along depth into $K$ blocks $\{B_i^{(k)}\}_{k=1}^{K}$ so that each block is a stack of $p$ layers  $B_i^{(k)}=F_i^{(l)} \circ F_i^{(l+1)}\dots \circ F_i^{(l+p)}$ and $k$ is its stage index. Inspired by the hierarchical property of deep neural networks, we aim to partition the neural networks according to their function level, for example, dividing the network into a ``low-level'' block that identifies curves and a ``high-level'' block that recognizes semantics. Although we cannot strictly differentiate ``low-level'' from ``high-level'', it is feasible to define  \emph{functional equivalence}.

\textbf{Definition 2} \textit{(Functional Equivalence) Given two functions $B$ and $B'$ with same input space $\mathcal{X}$ and output space $\mathcal{Y}$. $d: \mathcal{Y} \times \mathcal{Y} \to \mathbb{R}$ is the metric defined on $\mathcal{Y}$. For all inputs $\mathbf{x} \in \mathcal{X}$, if the outputs are the equivalent $d(B(\mathbf{x}),B'(\mathbf{x}))=0$, we say $B$ and $B'$ are  functional equivalent.}

A function is then uniquely determined by its peers who generate the same output with the same input. However, we can no longer define functional equivalence among neural networks, since network blocks might have varied input-output dimensions. It is neither possible to feed the same input to intermediate blocks with  different input dimensions, nor allow for a mathematically valid definition for metric space~\cite{elements1992,buldygin2000metric} when the output dimensions are not identical. We therefore resort to recent measurements on neural representation similarity~\cite{hardoon2004canonical, kornblith2019similarity} and define the functional similarity for neural networks. The intuition is simple: two networks are functionally similar when they produces similar outputs with similar inputs.

\textbf{Definition 3\label{def:3}} \textit{(Functional Similarity for Neural Networks) Assume we have a neural similarity index $s(\cdot,\cdot)$ and two neural networks $B: \mathcal{X} \in \mathbb{R}^{n \times d_{in}}\to \mathcal{Y} \in \mathbb{R}^{n \times d_{out}}$ and $B':\mathcal{X}' \in \mathbb{R}^{n \times d'_{in}}\to \mathcal{Y}' \in \mathbb{R}^{n \times d'_{out}}$. For any two batches of inputs $\mathbf{X} \subseteq \mathcal{X}$ and $\mathbf{X}' \subseteq \mathcal{X}'$ with large similarity $s(\mathbf{X}, \mathbf{X}')>\epsilon$, the functional similarity between $B$ and $B'$ are defined as their output similarity $s(B(\mathbf{X}),B'(\mathbf{X}'))$.}

This definition generalizes well to typical knowledge distillation~(KD)~\cite{hinton2015distilling} when $d_{in}=d'_{in}$, which we will elaborate in the Appendix. We also show in Appendix that Def.3 provides a necessary and insufficient condition for two identical networks. Using the method of Lagrange multipliers, the conditional similarity in Def.3 can be further simplified to  $S(B,B') = s(B(\mathbf{X}), B'(\mathbf{X}'))+ s(\mathbf{X}, \mathbf{X}')$, which is a summation of its input-output similarity. The full derivation is shown in the Appendix. 

\textbf{Finding the Equivalence Sets of Neural Blocks.}
With Def.3, we are equipped with the math tools to partition the networks into equivalent sets of blocks. Blocks in each set are expected to have high similarity, which are treated to be functionally equivalent and hence interchangeable.

With a graphical notion, we represent each neural network as a path \typo{graph}~$G(V,E)$~\cite{gross2018graph} with two nodes of vertex degree 1, and the other $n-2$ nodes of vertex degree 2. The ultimate goal is to find the best partition of each graph into $K$ disjoint sub-graphs along the depth, that sub-graph within each group has maximum internal functional similarity $S(B,B')$. In addition, we take a mild assumption that each sub-graph should have approximately similar size $|B_i^{(k)}| < (1+\epsilon)\frac{|\mathcal{M}_i|}{K}$, where $|\cdot|$ indicates the model size and $\epsilon$ is coefficient controls size limit for each block. We solve the above problem by posing a tri-level constrained optimization
{\footnotesize
\begin{align}
         \max_{B_{a_j}}  \quad&  J(A, \{B_i^{(k)}\}) = \max_{A_{(ik,p)} \in \{0,1\}} \sum_{i=1}^N \sum_{j=1}^{K} \sum_{k=1}^K A_{(ik,j)} S(B_i^{(k)*},B_{a_j}) & (\text{Clustering})\\
        s.t. \quad&\sum_{j=1}^{N_g} A_{(ik,j)}=1, \quad
        \{B_i^{(k)*}\}_{k=1}^K = \arg\max_{B_i^{(k)}} \sum_{k=1}^K A_{(ik,j)}S(B_i^{(k)},B_{a_j}) &(\text{Partition})\\
        & s.t.\quad 
        B_i^{(1)}\circ B_i^{(2)}\dots\circ B_i^{(K)}  = \mathcal{M}_i, B_i^{(k_1)} \cap B_j^{(k_2)} = \emptyset ,\forall k_1\neq k_2\\
        & \quad\quad |B_i^{(k)}| < (1+\epsilon)\frac{|\mathcal{M}_i|}{K}, k=1,\dots K
    \end{align}}
Where $A\in \mathbb{N}^{KN \times K}$ is the assignment matrix, where $A_{(ik,j)}=1$ denote the $B_{i}^{(k)}$ block belongs to the $j$-th equivalence set, otherwise 0. Note that each block only belongs to one equivalence set, thus each column sums up to 1, $\sum_{j=1}^{K} A_{(ik,j)}=1$. $B_{a_j}$ is the anchor node for the $j$-th equivalence set, which has the maximum summed similarly with all blocks in set $j$. 

The inner optimization largely resembles the conventional \emph{set cover problem}~\cite{hochba1997approximation} or $(K,1+\epsilon)$ \emph{graph partition problem}~\cite{karypis2000multilevel} that directly partition a graph into $k$ sets. Although the graph partition falls exactly in a NP-hard~\cite{hartmanis1982computers} problem, heuristic graph partitioning algorithms like Kernighan-Lin~(KL) algorithm~\cite{kernighan1970efficient} and Fiduccia–Mattheyses~(FM) algorithm~\cite{fiduccia1982linear} can be applied to solve our problem efficiently. In our implementation, we utilize a variant KL algorithm. With a random initialized network partition $\{B^{(k)}\}_{k=1}^K|_{t=0}$ for $\mathcal{M}$ at $t=0$, we iteratively find the optimal separation by swapping nodes~(network layer). Given the two consecutive block $B^{(k)}|_{t}=F_i^{(l)}  \dots \circ  F_i^{(l+p_{k})}$ and $B^{(k+1)}|_{t}=F_i^{(l+p_{k}+1)} \dots \circ F_i^{(l+p_{k}+p_{k+1})}$ at time $t$, 
we conduct a forward and a backward neural network layer swap between successive blocks, whereas the partition achieving the largest objective value becomes the new partition
{\footnotesize
\begin{align}
        (B^{(k)}|_{t+1},B^{(k+1)}|_{t+1})&=\arg\max \{J(B^{(k)}|_{t},B_{i}^{(k+1)}|_{t}), J(B_{i}^{(k)}|_t^{\text{f}},B_{i}^{(k+1)}|_t^{\text{f}}),J(B_{i}^{(k)}|_{t}^{\text{b}},B_{i}^{(k+1)}|_{t}^{\text{b}})\}\\
    \text{where }
    (B_{i}^{(k)}|_t^{\text{f}},B_{i}^{(k+1)}|_t^{\text{f}}) & = B^{(k)}|_{t} \xrightarrow[]{F_i^{l+p_k}} B_{i}^{(k+1)} , (B_{i}^{(k)}|_t^{\text{b}},B_{i}^{(k+1)}|_t^{\text{b}}) = B^{(k)}|_{t}\xleftarrow[]{\tiny{F_i^{l+p_k+1}}} B_{i}^{(k+1)}
\end{align}}
For the outer optimization, we do a  K-Means~\cite{macqueen1967some} style clustering. With the current network partition $\{B^{(k)*}\}_{k=1}^K$, we alternate between assigning each block to a equivalence set $G_j$, and identifying the anchor block within each set $B_{a_j} \in G_j$. It has been proved that both KL and K-Means algorithms converge to a local minimum according to the initial partition and anchor selection. We repeat the optimization for $R=200$ runs with different seeds and select the best partition as our final results.

\subsection{Network Reassembly by Solving an Integer Program}
As we have divided each deep network into $K$ partitions, each belongs to one of the $K$ equivalence sets, all we want now is to find the best combination of neural blocks as a new pre-trained model under certain computational constraints. Consider $K$ disjoint equivalence sets $G_1, \dots ,G_{K}$ of blocks to be reassembled into a new deep network of parameter constraint $C_{\text{param}}$ and computational constraint $C_{\text{FLOPs}}$, the objective is to choose exactly one block from each group $G_j$ as well as from each network stage index $j$ such that the reassembled model achieves optimal performance on the 
target task without exceeding the capacity. We introduce two the binary matrices $X_{(ik,j)}$ and $Y_{(ik,j)}$ to uniquely identity the reassembled model $\mathcal{M}(X, Y)$. $X_{(ik,j)}$ takes on value 1 if and only if $B_{i}^{(k)}$ is chosen in group $G_j$, and $Y_{(ik,j)}=1$ if $B_{i}^{(k)}$ comes from the $k$-th block. The selected blocks are arranged by the block stage index. The problem is formulated as 
{\footnotesize\begin{align}
    \max_{X, Y} & P_{T}(\mathcal{M}(X, Y)) \label{eq:reasemble} \\
    \text{s.t. }& |\mathcal{M}(X, Y)| \leq C_{\text{param}},
    FLOPs(\mathcal{M}(X, Y)) \leq C_{\text{FLOPs}} \\
    & \sum_{i=1}^N \sum_{k=1}^K X_{(ik,j)}=1, X_{(ik,j)} \in \{0,1\}, j=1,\dots, K \label{eq:only group}\\
    & \sum_{i=1}^N \sum_{j=1}^K Y_{(ik,j)} = 1, Y_{(ik,j)} \in \{0,1\}, k=1,\dots, K \label{eq:only depth}
\end{align}}
where $P_{T}$ is again the task performance. Equation~\ref{eq:only group} and~\ref{eq:only depth} indicates that each model only \typo{possesses} a single block from each equivalence set and each stage index. It falls exactly into a \textit{0-1 Integer Programming}~\cite{papadimitriou1998combinatorial} problem with a non-linear objective. Conventional methods train each $\mathcal{M}(X,Y)$ to obtain $P_{T}$. Instead of training each candidate till convergence, we estimate the transfer-ability of a network by counting the linear regions in the network as a training-free proxy. 

\textbf{Estimating the Performance with Training-Free Proxy.} 
The number of linear region~\cite{montufar2014number,hanin2019complexity} is a theoretical-grounded tool to describe the expressivity of a neural network, which has been successfully applied on NAS without training~\cite{mellor2021neural, chen2020tenas}. We, therefore, calculate the data-dependent linear region to estimate the transfer performance of each model-task combination. The intuition is straightforward: the network can hardly learn to distinguish inputs with similar binary codes.

We apply random search to get a generation of reassembly candidates. For a whole mini-batch of inputs, we feed them into each network and binarilize the features vectors using a sign function. Similar to NASWOT~\cite{mellor2021neural}, we compute the kernel matrix $\mathbf{K}$ using Hamming distance $d(\cdot, \cdot)$ and rank the models using $\log(\det \mathbf{K})$. Since the computation of $\mathbf{K}$ requires nothing more than a few batches of network forwarding, we replace $P_{T}$ in Equation~\ref{eq:reasemble} with NASWOT score for fast model evaluation.

\section{Experiments}
In this section, we first explore some basic properties of the the proposed \name~task, and then evaluate our solution on a series of transfer learning benchmarks to verify its efficiency. 
\label{sec:exp}

\textbf{Model Zoo Setup.} We construct our model zoo by collecting pre-trained weights from Torchvision~\footnote{https://pytorch.org/vision/stable/index.html}, \texttt{timm}~\footnote{https://github.com/rwightman/pytorch-image-models} and OpenMMlab~\footnote{https://github.com/open-mmlab}. We includes a series of manually designed CNN models like ResNet~\cite{he2016deep} and ResNeXt\cite{xie2017aggregated}, as well as NAS-based architectures like  RegNetY~\cite{radosavovic2020designing} and MobileNetv3~\cite{howard2019searching}. Due to recent popularity of vision transformer, we also take several well-known attention-based \typo{architectures} into consideration, including Vision Transformer (ViT)~\cite{dosovitskiy2020image} and Swin-Transformer~\cite{liu2021Swin}. In addition to the differentiation of the network structure pre-trained on ImageNet, we include models with a variety of pre-trained strategies, including SimCLR~\cite{chen2020simple}, MoCov2~\cite{chen2020improved} and BYOL~\cite{grill2020bootstrap} for ResNet50, MoCov3~\cite{chen2021empirical} and MAE~\cite{MaskedAutoencoders2021} for ViT-B. Those models are pre-trained on ImageNet1k~\cite{russakovsky2015imagenet}, ImageNet21K~\cite{ridnik2021imagenet21k}, Xrays~\cite{Cohen2022xrv} and iNaturalist2021~\cite{van2021benchmarking}, Finally we result in 21 network architectures, with 30 pre-trained weights in total. We manually identify the atomic node to satisfy our line graph assumption. Each network is therefore a 
line graph composed of atomic nodes. 

\textbf{Implementation details.} For all experiments, we set the partition number $K=4$ and the block size coefficient $\epsilon=0.2$. We sample 1/20 samples from each train set to calculate the linear CKA representation similarity. The NASWOT~\cite{mellor2021neural} score is estimated with 5-batch average, where each mini-batch contains 32 samples. We set 5 levels of computational constraints, with $C_{\text{param}} \in \{10,20,30,50,90\}$ and  $C_{\text{FLOPs}} \in \{3,5,6,10,20\}$, which is denoted as \name($K,C_{\text{param}}$,$C_{\text{FLOPs}}$). For each setting, we randomly generated 500 candidates. Each reassembled model is evaluate under 2 protocols (1) \textsc{frozen-tuning.} We freeze all trained blocks and only update the parameter for the stitching layer and the last linear classifier and (2) \textsc{Full-turning.} All network parameter are updated. All experiments are conducted on a $8\times$GeForce RTX 3090 server. \typo{To reduce the feature similarity calculation cost, we construct the similarity table \emph{offline} on ImageNet. The complexity analysis and full derivation are shown in the Appendix. }

\subsection{Exploring the Properties for Deep Reassembly}

\begin{figure}
\begin{minipage}{.51\linewidth}
    \centering
    \includegraphics[width=\linewidth]{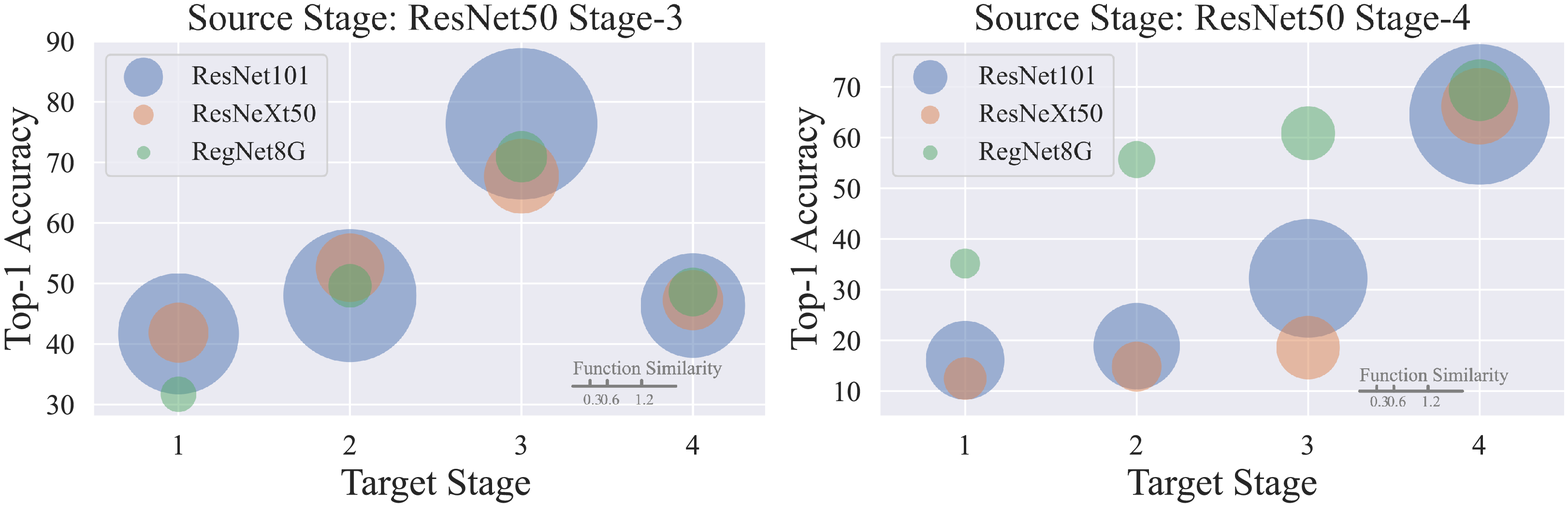}
    \caption{\textsc{frozen-tuning} accuracy on ImageNet by replacing the 3$^{\text{nd}}$ and 4$^{\text{th}}$ stage of R50 to target blocks. }
    \label{fig:similarity and peformance}
\end{minipage}
\hfill
\begin{minipage}{.48\linewidth}
    \centering
    \includegraphics[width=\linewidth]{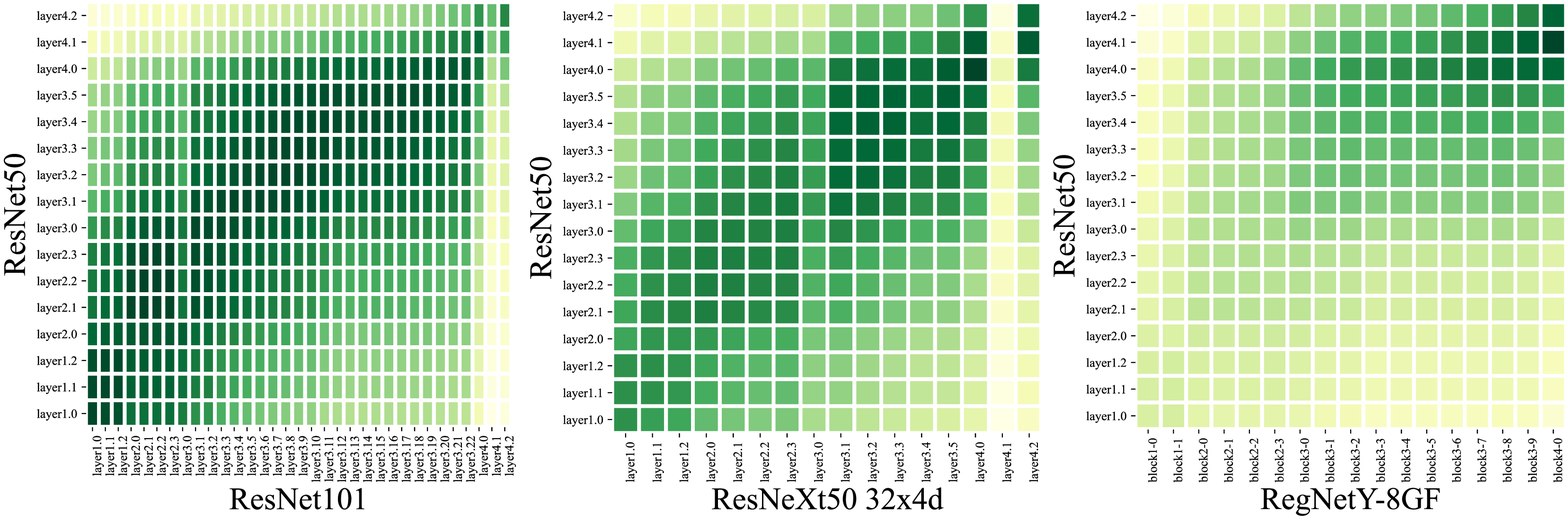}
    \caption{Pair-wise Linear CKA between pre-trained R50 and (1) R101 (2) RX50 and (3) Reg8G.}
    \label{fig:cka2}
\end{minipage}
    \vspace{-4mm}
\end{figure}

\textbf{Similarity, Position and Reassembly-ability.} Figure~\ref{fig:similarity and peformance} validates our functional similarity, reassembled block selection, and its effect on the model performance. For the ResNet50 trained on \typo{ImageNet}, we replace its 3$^{\text{nd}}$ and 4$^{\text{th}}$ stage with a target block from another pre-trained network~(ResNet101, ResNeXt50 and RegNetY8G), connected by a single stitching layer. Then, the reassembled networks are re-trained on ImageNet for a 20 epochs under \textsc{frozen-turning} protocol. The derived functional similarity in Section~\ref{sec:funcsim} is shown as the diameter of each circle. \textbf{We observe that}, the stitching position makes a substantial difference regarding the reassembled model performance. When replaced with a target block with the same stage index, the reassembled model performs surprisingly well, with $\geq 70\%$ top-1 accuracy, even if its predecessors are trained with different architectures, seeds, and hyperparameters. \textbf{It is also noted} that, though function similarity is not numerically proportional to the target performance, it correctly reflects the performance ranking within the same target network. It suggests that our function similarity provides a reasonable criteria to identify equivalence set. In sum, the coupling between the \emph{similarity-position-performance} explains our design to select one block from each equivalence set as well as the stage index. We also visualize the linear CKA~\cite{kornblith2019similarity} similarity between the R50 and the target networks in Figure~\ref{fig:cka2}. An interesting finding is that \emph{diagonal pattern} for the feature similarity. The representation at the same stage is highly similar. More similarity visualizations are provided in the Appendix.

\textbf{Partition Results.} Due to the space limitation, the partition results of the model zoo are provided in the Appendix. Our observation is that, the equivalent sets tend to \emph{cluster the blocks by stage index}. For example, all bottom layers of varied pre-trained networks are within the same equivalence set. It provides valuable insight that neural networks learns similar patterns at similar network stage.

\textbf{Architecture or Pre-trained Weight.} Since \name~searches for the architecture and weights concurrently, a natural question arises that ``Do both \emph{architecture} and \emph{pre-trained weights} lead to the final improvement? Or only \emph{architecture} counts?'' We provide the experiments in the Appendix that both factors contribute. It is observed that training the \name~architecture from scratch leads to a substantial performance drop compared with \name~model with both new structures and pre-trained weights. It validates our arguments that our reassembled models benefit from the pre-trained models for efficient transfer learning.  

\begin{figure}
    \centering
    \includegraphics[width=\linewidth]{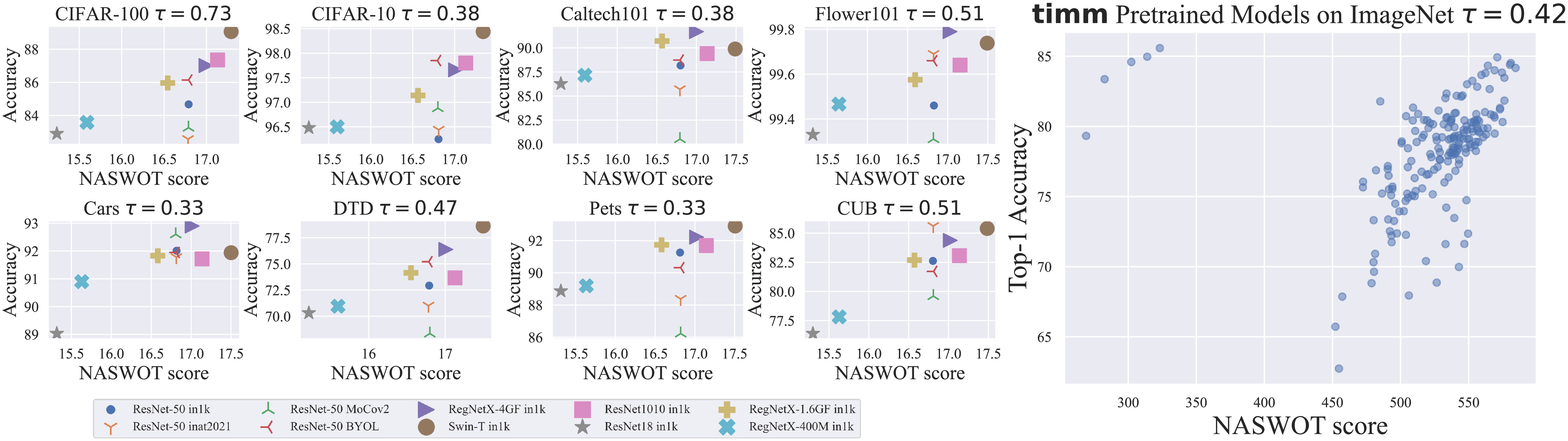}
    \caption{Plots of NASWOT~\cite{mellor2021neural} score and test accuracy for (Left) 10 pre-trained model on 8 downstream tasks and (Right)  \texttt{timm} model zoo on ImageNet. $\tau$ is the Kendall’s Tau correlation.}
    \label{fig:linsep_for_transfer}
\end{figure}

\textbf{Verifying the training-free proxy.} As the first attempt to apply the NASWOT to measure model transfer-ability, we verify its efficacy before applying it to \name~task. We adopt the score to rank 10 pre-trained models on 8 image classification tasks, as well as the \texttt{timm} model zoo on ImageNet, shown in Figure~\ref{fig:linsep_for_transfer}. We also compute the Kendall’s Tau correlation~\cite{kendall1938new} between the fine-tuned accuracy and the NASWOT score. It is observed that the NASWOT score provides a reasonable predictor for model transfer-ability with a high Kendall’s Tau correlation. 

\subsection{Transfer learning with Reassembled Model}
\begin{figure}
\renewcommand{\arraystretch}{1.1}
\setlength{\tabcolsep}{3pt}
\centering
\begin{minipage}[b]{.49\linewidth}
    \centering
    \scriptsize
    \begin{tabular}{l|r|r|c}
    \toprule
      Architecture  & \#Train/All Params (M) & FLOPs (G) & Top-1 \\
      \hline
      RSB-ResNet-18  & 11.69/11.69 &	1.82 & 70.6 \\
        RegNetY-800M &  6.30/6.30
        & 0.8 & \underline{76.3}\\
         ViT-T16 &  5.7/5.7 
        & 1.3 & 74.1\\
        \hline
        \name(4,10,3)-FZ$^{\dag}$ & \textcolor{Red}{1.02}/7.83 & 2.99 & 41.2 \\
        \rowcolor{gray!10}
        \name(4,10,3)-FT$^{\dag}$ &  7.83/7.83 & 2.99 & 76.9\\
        \rowcolor{gray!15}
        \name(4,10,3)-FT &  7.83/7.83 & 2.99 & 78.4\\
        \hline\hline
         RSB-ResNet-50  & 25.56/25.56&	4.12 & 79.8\\
        RegNetY-4GF & 20.60/20.60 & 4.0 & 79.4\\
         ViT-S16&  22.0/22.0  & 4.6 & 79.6\\
         Swin-T & 28.29/28.29 &	4.36 & \underline{81.2}\\\hline
         \name(4,30,6)-FZ$^{\dag}$ &  \textcolor{Red}{1.57}/24.89  & 4.47  & 60.5\\
         \rowcolor{gray!10}
        \name(4,30,6)-FT$^{\dag}$ & 24.89/24.89  & 4.47  &  79.6\\
        \rowcolor{gray!15}
        \name(4,30,6)-FT &  24.89/24.89  & 4.47  & 81.2 \\
        \hline\hline
        RSB-ResNet-101  &  44.55/44.55 & 7.85 & 81.3\\
        RegNetY-8GF &  39.20/39.20 & 8.1 & 81.7\\
        Swin-S &  49.61/49.61	 & 8.52& \underline{82.8}\\
        \hline
        \name(4,50,10)-FZ$^{\dag}$ & \textcolor{Red}{3.92}/40.41  & 6.43  & 72.0\\
        \rowcolor{gray!10}
        \name(4,50,10)-FT$^{\dag}$ &  40.41/40.41  &  6.43  & 81.3 \\
        \rowcolor{gray!15}
        \name(4,50,10)-FT &  40.41/40.41  & 6.43  & 82.3 \\
        \hline\hline
        RegNetY-16GF & 83.6/83.6 & 16.0 & 82.9\\
        ViT-B16&86.86/86.86 &	33.03 & 79.8\\
        Swin-B & 87.77/87.77	& 15.14& \underline{83.1}\\
        \hline
        \name(4,90,20)-FZ$^{\dag}$ & \textcolor{Red}{1.27}/ 80.66  & 13.29  & 78.6\\
        \rowcolor{gray!10}
        \name(4,90,20)-FT$^{\dag}$ &  80.66/ 80.66  &  13.29  & 82.4 \\
        \rowcolor{gray!15}
        \name(4,90,20)-FT & 80.66/ 80.66  & 13.29  & 83.2 \\
        \bottomrule
    \end{tabular}
     \renewcommand\figurename{Table}
    \caption{Top-1 accuracy of models trained on ImageNet. \dag~means the model is trained for 100 epochs. ``FZ'' and ``FT'' denote the reassembled blocks are frozen or fine-tuned. Trainable parameters are marked in \textcolor{Red}{red}.}
    \label{tab:imagenet}
\end{minipage}
\hfill
\begin{minipage}[b]{.49\linewidth}
   \begin{center}
    \includegraphics[width=0.7\textwidth]{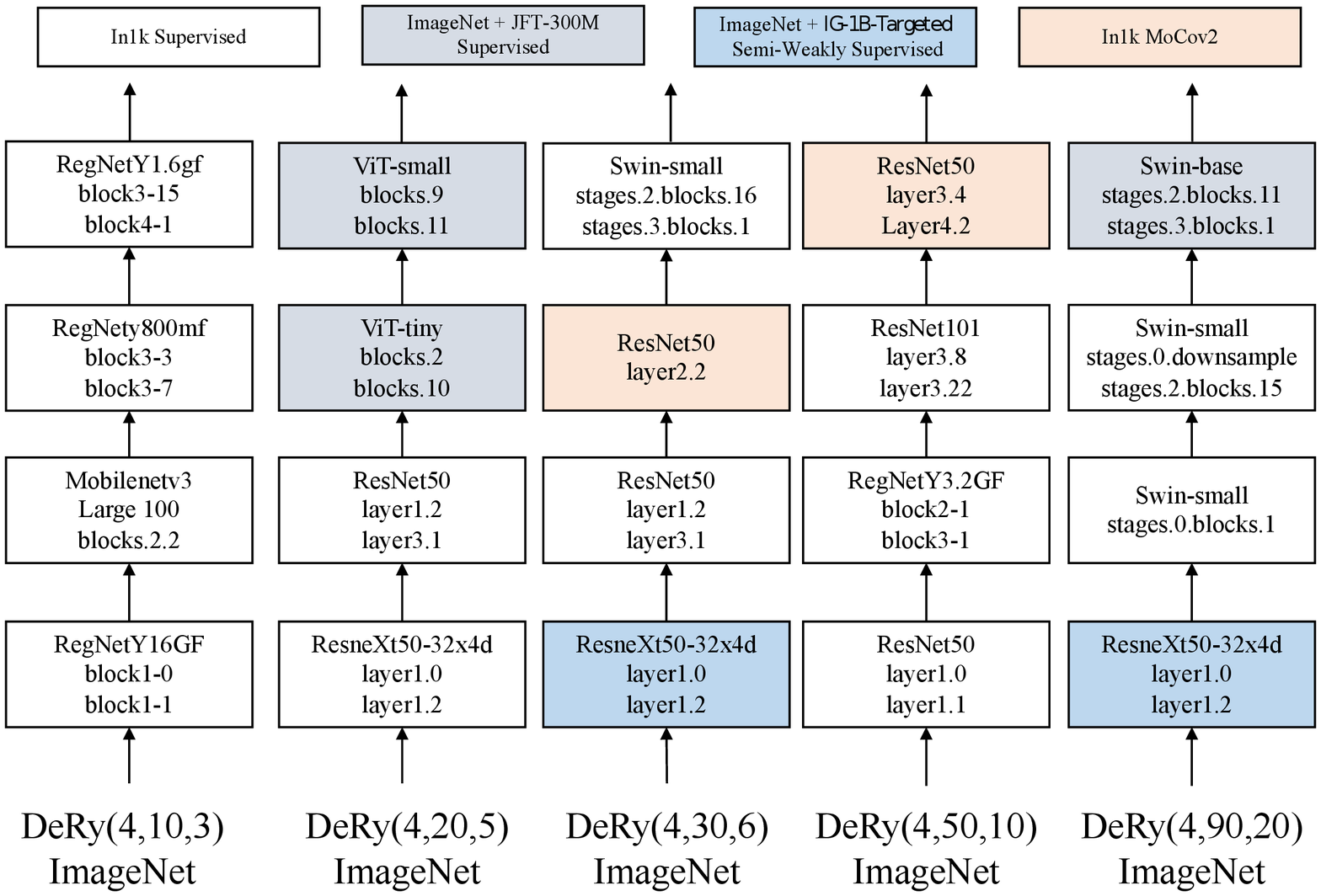}
  \end{center}
  \vspace{-4mm}
  \caption{Reassembled structures on ImageNet.}
  \label{fig:reassemled_arch}
 \renewcommand{\arraystretch}{0.8}
\setlength{\tabcolsep}{1pt}
    \centering
    \scriptsize
        \begin{tabular}{l|r|r|c|c}
    \toprule
       Architecture  & Params (M) & FLOPs (G) & Top-1 & Top-5 \\
       \hline
        ResNet-50  &  25.56&	4.12 & 76.8 & 93.3\\
        Swin-T  &  28.29 & 4.36 & 78.3&	94.6\\
       \rowcolor{gray!15}
        \name(30, 6)-FT & \textbf{24.89} & 4.47 & \textbf{79.6} & \textbf{94.8}\\\hline
        ResNet-101  &  44.55 & 7.85 & 79.0 & 94.5\\
        Swin-S  &  49.61 &8.52 & 80.8	& \textbf{95.7}\\
        \rowcolor{gray!15}
        \name(50, 10)-FT & \textbf{40.41}  & \textbf{6.43}  &  \textbf{81.2} & 95.6
       \\\bottomrule
    \end{tabular}
      \renewcommand\figurename{Table}
      \vspace{-2mm}
    \caption{Top-1 and Top-5 Accuracy for the ImageNet 100-epoch \textsc{Full-tuning} experiment.}
    \label{tab:short-imagenet}
\centering
     \includegraphics[width=0.7\linewidth]{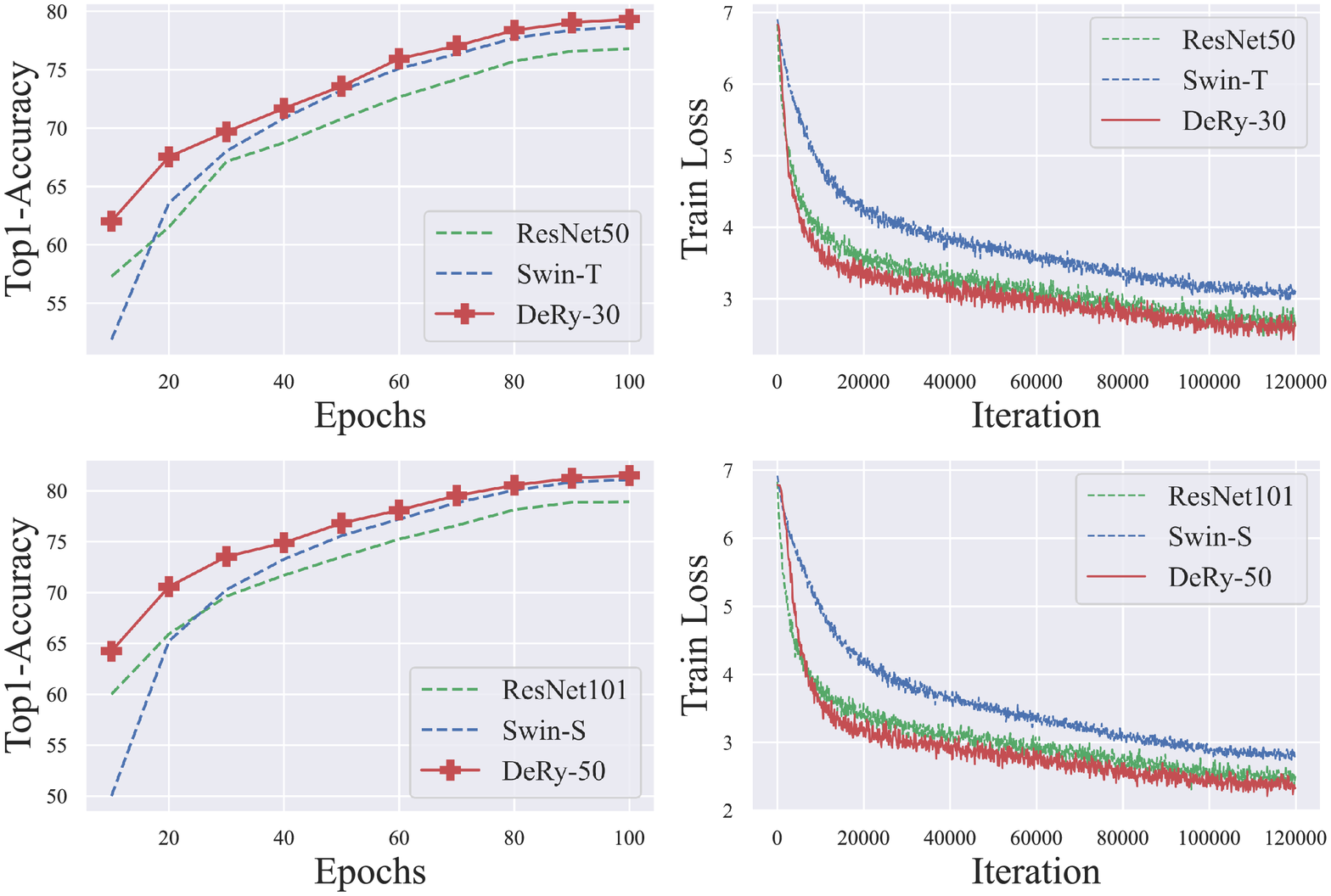}
     \vspace{-2mm}
    \caption{(Left) Test accuracy and (Right) Train loss comparison under the 100-epoch training on ImageNet.}
    \label{fig:short-imagenet}
\end{minipage}
\end{figure}

\textbf{Evaluation on ImageNet1k.} We first compare the reassembled network on ImageNet~\cite{russakovsky2015imagenet} with current best-performed architectures. We train each model for either 100 epochs as \textsc{Short-training} or a 300 epochs as \textsc{full-training}. Except for \name, all models are trained from scratch. We optimize each network with AdamW~\cite{loshchilov2017decoupled} alongside a initial learning rate of $1e-3$ and cosine lr-decay, mini-batch of 1024 and weight decay of 0.05. We apply RandAug~\cite{cubuk2020randaugment}, Mixup~\cite{zhang2017mixup} and CutMix~\cite{yun2019cutmix} as data augmentation. All model are trained and tested on $224$ image resolutions.

Table~\ref{tab:imagenet} provides the Top-1 accuracy comparison on Imagenet with various computational constraint. We \underline{underline} the best-performed model in the model zoo. \textbf{First}, It is worth-noting that \name~provide very competitive model, even under \textsc{frozen-turning} or \textsc{Short-training} protocol. \name(4,90,20) manages to reach $78.6\%$ with 1.27M parameter trainable, which provides convincing clue that the heterogeneous trained model are largely graftable. With only \textsc{Short-training}, \name~models also match up with the full-time trained model in the zoo. For example, \name(4,10,3) gets to $76.9\%$ accuracy within 100 epochs' training, surpassing all small-sized models. The performance can be further improved towards $78.4\%$ with the standard 300-epoch training. \textbf{Second}, \name~brings about faster convergence. We compare with ResNet-50 and Swin-T under the same \textsc{Short-training} setting in Table~\ref{tab:short-imagenet} and Figure~\ref{fig:short-imagenet}. It is clear that, by assembling the off-the-self pre-trained blocks, the \name~models can be optimized faster than the it competitors, achieving 0.9\% and 0.2\% accuracy improvement over the Swin-T model with less parameter and computational requirements. \textbf{Third}, as showcased in Figure~\ref{fig:reassemled_arch}, our \name~is able to search for diverse and hybrid network structures. \name(4,10,3) learns to adopt light-weight blocks like MobileNetv3, while \name(4,90,20) gets to a large CNN-Swin hybrid architecture. Similar hybrid strategy has been proved to be efficient in manual network design~\cite{mehta2021mobilevit, xiao2021early}.

\textbf{Transfer Image classification.} We evaluate transfer learning performance on 9 natural image datasets. These datasets covered a wide range of image classification tasks, including 3 object classification tasks CIFAR-10~\cite{krizhevsky2009learning}, CIFAR-100~\cite{krizhevsky2009learning} and Caltech-101~\cite{fei2004learning}; 5 fine-grained classification tasks Flower-102~\cite{nilsback2008automated}, Stanford Cars~\cite{KrauseStarkDengFei-Fei_3DRR2013}, FGVC Aircraft\cite{maji13fine-grained}, Oxford-IIIT Pets~\cite{parkhi12a} and CUB-Bird~\cite{cubbird} and 1 \typo{texture} classification task DTD~\cite{cimpoi14describing}. We \textsc{full-tune} all candidate networks in the model zoo and compare them with our \name~model. Two model selection strategies LogME~\cite{you2021logme} and LEEP~\cite{nguyen2020leep} are also taken as our baselines. For fair comparison, we further train the reassembled network on ImageNet for 100 epochs to further boost the transfer performance. Following \cite{chen2020simple, kornblith2019better}, we perform hyperparameter
tuning for each model-task combination, which are elaborated in the Appendix. 

Figure~\ref{fig:transfer} compares the transfer performance between our proposed \name~and all candidate models. By constructing models from building blocks, the \name~generally surpasses all network trained from scratch within the same computational constraints, even beats pre-trained ones on Cars, Aircraft, and Flower. If allowing for pre-training on ImageNet~(\name+In1k), we can further promote the test accuracy, even better than the best-performing candidate in the original model zoo~(highlighted by $\times$). The performance improvement rises up as parameter constraints increase, which demonstrates the scalability of the proposed solution. Model selection approaches like LogME and LEEP may not necessarily get the optimal model, thus failing to release the full potential of the model zoo. These findings provide encouraging evidence that \name~gives rise to an alternative approach to improve the performance when transferring from a zoo of models.


\begin{figure}
    \centering
    \includegraphics[width=\linewidth]{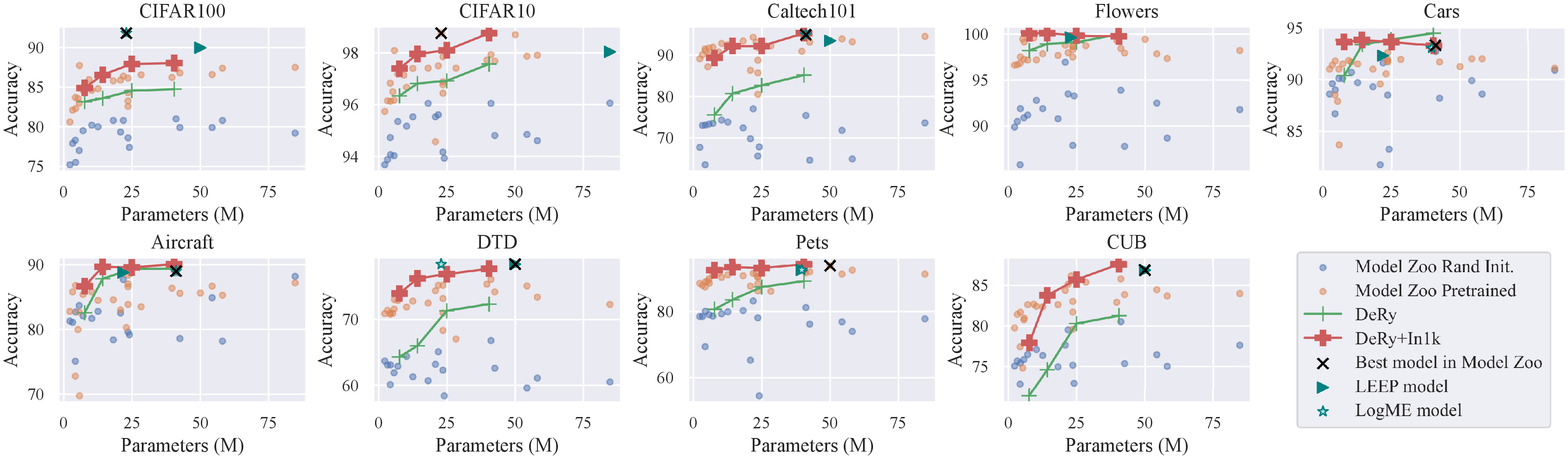}
    \vspace{-4mm}
    \caption{Transfer performance on 9 image classification tasks with the model zoo and our \name. Each 
\textcolor{NavyBlue}{blue} or \textcolor{BurntOrange}{orange} point refers to a single model trained from scratch or pre-trained weights.}
    \vspace{-4mm}
    \label{fig:transfer}
\end{figure}

\begin{wraptable}{r}{5.5cm}
\vspace{-5mm}
\scriptsize
    \begin{tabular}{c|c |c |c}
    \toprule
         \pbox{1.4cm}{\centering Cover Set Partition} & \pbox{1.6cm}{\centering Train-Free
         Reassembly} & Acc~(\%) & \pbox{1.4cm}{\centering Search
         Cost
         (GPU days)} \\
         \hline
         \textcolor{Green}{\Checkmark} & \textcolor{Green}{\Checkmark} & 72.0 & 0.23\\
         \hline
         \textcolor{red}{\XSolidBrush} & \textcolor{Green}{\Checkmark}  & 70.5 & 1.48 \\
         \hline
         \textcolor{Green}{\Checkmark} & \textcolor{red}{\XSolidBrush} & 73.5 & 135 \\\hline
         \textcolor{red}{\XSolidBrush} & \textcolor{red}{\XSolidBrush} & 72.2 & 135 \\
         \bottomrule
    \end{tabular}
    \vspace{-2mm}
    \caption{Ablation study on partition and reassembly strategy.}
    \label{tab:ablation}
\vspace{-3mm}
\end{wraptable} 
\textbf{Ablation Study.} To study the influence of each stage in our solution, we conduct ablation study by  replacing the (1) cover set partition and (2) training-free reassembly with a random search individually. For the partition ablation, we randomly dissect each network into $K$ partitions and reassemble the blocks in an order-less manner using our training-free proxy. For the reassembly ablation, we retain the cover set partition and fine-tune each randomly reassembled network for 100 epochs. Due to the computation limitation, we can only 
evaluate 25 candidates for reassembly ablation. We report the 100-epoch~\textsc{frozen-tuning} top-1 accuracy and the search time on ImageNet in Table~\ref{tab:ablation} under the \name(4,50,10) setting. \typo{Note that we do not include the similarity computation time into our account since it is computed \emph{offline}.} We see that the majority of the search cost comes from the fine-tuning stage. The training-free proxy largely alleviates the tremendous computational cost by $10^4$ times, with marginal performance degradation. On the other hand, the cover set model partition not only improves the transfer performance but also reduces the reassembly search space from $O(\prod_{i=1}^N \binom{L_i-1}{K-1})$ to $O(1)$. Both stages are crucial.



\section{Conclusion}
In this study, we explore a novel knowledge-transfer task called Deep Model Reassembly~(\name). \name~seeks to deconstruct heterogeneous pre-trained neural networks into building blocks and then reassemble them into models subject to user-defined constraints. We provide a proof-of-concept solution to show that \name ~can be made not only possible but practically efficient.
 Specifically, pre-trained networks are partitioned jointly via a cover set optimization to form a series of equivalence sets. The learned equivalence sets enable choosing and assembling blocks to customize networks, which is accomplished by solving integer program with a training-free task-performance proxy. \name~not only achieves gratifying performance on a series of transfer learning benchmarks, but also sheds light on the functional similarity between neural networks by stitching heterogeneous models.
 
\vspace{3mm}
\textbf{Acknowledgement}

This research is supported by the National Research Foundation Singapore under its AI Singapore Programme (Award Number: AISG2-RP-2021-023).
Xinchao Wang is the corresponding author.

{\small
\bibliographystyle{plain}
\bibliography{egbib}}

\begin{thebibliography}{10}

\bibitem{agostinelli2021transferability}
Andrea Agostinelli, Jasper Uijlings, Thomas Mensink, and Vittorio Ferrari.
\newblock Transferability metrics for selecting source model ensembles.
\newblock {\em arXiv preprint arXiv:2111.13011}, 2021.

\bibitem{bansal2021revisiting}
Yamini Bansal, Preetum Nakkiran, and Boaz Barak.
\newblock Revisiting model stitching to compare neural representations.
\newblock {\em Advances in Neural Information Processing Systems}, 34, 2021.

\bibitem{bao2019information}
Yajie Bao, Yang Li, Shao-Lun Huang, Lin Zhang, Lizhong Zheng, Amir Zamir, and
  Leonidas Guibas.
\newblock An information-theoretic approach to transferability in task transfer
  learning.
\newblock In {\em 2019 IEEE International Conference on Image Processing
  (ICIP)}, pages 2309--2313. IEEE, 2019.

\bibitem{bolya2021scalable}
Daniel Bolya, Rohit Mittapalli, and Judy Hoffman.
\newblock Scalable diverse model selection for accessible transfer learning.
\newblock {\em Advances in Neural Information Processing Systems}, 34, 2021.

\bibitem{buldygin2000metric}
Valeri\u\i~Vladimirovich Buldygin and IU~V Kozachenko.
\newblock {\em Metric characterization of random variables and random
  processes}, volume 188.
\newblock American Mathematical Soc., 2000.

\bibitem{chen2020simple}
Ting Chen, Simon Kornblith, Mohammad Norouzi, and Geoffrey Hinton.
\newblock A simple framework for contrastive learning of visual
  representations.
\newblock In {\em International conference on machine learning}, pages
  1597--1607. PMLR, 2020.

\bibitem{chen2020tenas}
Wuyang Chen, Xinyu Gong, and Zhangyang Wang.
\newblock Neural architecture search on imagenet in four gpu hours: A
  theoretically inspired perspective.
\newblock In {\em International Conference on Learning Representations}, 2021.

\bibitem{chen2020improved}
Xinlei Chen, Haoqi Fan, Ross Girshick, and Kaiming He.
\newblock Improved baselines with momentum contrastive learning.
\newblock {\em arXiv preprint arXiv:2003.04297}, 2020.

\bibitem{chen2021empirical}
Xinlei Chen, Saining Xie, and Kaiming He.
\newblock An empirical study of training self-supervised vision transformers.
\newblock In {\em Proceedings of the IEEE/CVF International Conference on
  Computer Vision}, pages 9640--9649, 2021.

\bibitem{elements1992}
B.~Choudhary.
\newblock {\em The Elements of Complex Analysis}.
\newblock New Age International Publishers, 1992.

\bibitem{cimpoi14describing}
M.~Cimpoi, S.~Maji, I.~Kokkinos, S.~Mohamed, , and A.~Vedaldi.
\newblock Describing textures in the wild.
\newblock In {\em Proceedings of the {IEEE} Conf. on Computer Vision and
  Pattern Recognition ({CVPR})}, 2014.

\bibitem{Cohen2022xrv}
Joseph~Paul Cohen, Joseph~D. Viviano, Paul Bertin, Paul Morrison, Parsa
  Torabian, Matteo Guarrera, Matthew~P Lungren, Akshay Chaudhari, Rupert
  Brooks, Mohammad Hashir, and Hadrien Bertrand.
\newblock {TorchXRayVision: A library of chest X-ray datasets and models}.
\newblock In {\em Medical Imaging with Deep Learning}, 2022.

\bibitem{csiszarik2021similarity}
Adri{\'a}n Csisz{\'a}rik, P{\'e}ter K{\H{o}}r{\"o}si-Szab{\'o}, {\'A}kos
  Matszangosz, Gergely Papp, and D{\'a}niel Varga.
\newblock Similarity and matching of neural network representations.
\newblock {\em Advances in Neural Information Processing Systems}, 34, 2021.

\bibitem{cubuk2020randaugment}
Ekin~D Cubuk, Barret Zoph, Jonathon Shlens, and Quoc~V Le.
\newblock Randaugment: Practical automated data augmentation with a reduced
  search space.
\newblock In {\em Proceedings of the IEEE/CVF Conference on Computer Vision and
  Pattern Recognition Workshops}, pages 702--703, 2020.

\bibitem{dai2011greedy}
Dong Dai and Tong Zhang.
\newblock Greedy model averaging.
\newblock {\em Advances in Neural Information Processing Systems}, 24, 2011.

\bibitem{dietterich2000ensemble}
Thomas~G Dietterich.
\newblock Ensemble methods in machine learning.
\newblock In {\em International workshop on multiple classifier systems}, pages
  1--15. Springer, 2000.

\bibitem{dosovitskiy2020image}
Alexey Dosovitskiy, Lucas Beyer, Alexander Kolesnikov, Dirk Weissenborn,
  Xiaohua Zhai, Thomas Unterthiner, Mostafa Dehghani, Matthias Minderer, Georg
  Heigold, Sylvain Gelly, et~al.
\newblock An image is worth 16x16 words: Transformers for image recognition at
  scale.
\newblock {\em arXiv preprint arXiv:2010.11929}, 2020.

\bibitem{feder1999complexity}
Tomas Feder, Pavol Hell, Sulamita Klein, and Rajeev Motwani.
\newblock Complexity of graph partition problems.
\newblock In {\em Proceedings of the thirty-first annual ACM symposium on
  Theory of computing}, pages 464--472, 1999.

\bibitem{fei2004learning}
Li~Fei-Fei, Rob Fergus, and Pietro Perona.
\newblock Learning generative visual models from few training examples: An
  incremental bayesian approach tested on 101 object categories.
\newblock In {\em 2004 conference on computer vision and pattern recognition
  workshop}, pages 178--178. IEEE, 2004.

\bibitem{fiduccia1982linear}
Charles~M Fiduccia and Robert~M Mattheyses.
\newblock A linear-time heuristic for improving network partitions.
\newblock In {\em 19th design automation conference}, pages 175--181. IEEE,
  1982.

\bibitem{grill2020bootstrap}
Jean-Bastien Grill, Florian Strub, Florent Altch{\'e}, Corentin Tallec, Pierre
  Richemond, Elena Buchatskaya, Carl Doersch, Bernardo Avila~Pires, Zhaohan
  Guo, Mohammad Gheshlaghi~Azar, et~al.
\newblock Bootstrap your own latent-a new approach to self-supervised learning.
\newblock {\em Advances in Neural Information Processing Systems},
  33:21271--21284, 2020.

\bibitem{gross2018graph}
Jonathan~L Gross, Jay Yellen, and Mark Anderson.
\newblock {\em Graph theory and its applications}.
\newblock Chapman and Hall/CRC, 2018.

\bibitem{hanin2019complexity}
Boris Hanin and David Rolnick.
\newblock Complexity of linear regions in deep networks.
\newblock In {\em International Conference on Machine Learning}, pages
  2596--2604. PMLR, 2019.

\bibitem{hardoon2004canonical}
David~R Hardoon, Sandor Szedmak, and John Shawe-Taylor.
\newblock Canonical correlation analysis: An overview with application to
  learning methods.
\newblock {\em Neural computation}, 16(12):2639--2664, 2004.

\bibitem{hartmanis1982computers}
Juris Hartmanis.
\newblock Computers and intractability: a guide to the theory of
  np-completeness (michael r. garey and david s. johnson).
\newblock {\em Siam Review}, 24(1):90, 1982.

\bibitem{MaskedAutoencoders2021}
Kaiming He, Xinlei Chen, Saining Xie, Yanghao Li, Piotr Doll{\'a}r, and Ross
  Girshick.
\newblock Masked autoencoders are scalable vision learners.
\newblock {\em arXiv:2111.06377}, 2021.

\bibitem{he2016deep}
Kaiming He, Xiangyu Zhang, Shaoqing Ren, and Jian Sun.
\newblock Deep residual learning for image recognition.
\newblock In {\em Proceedings of the IEEE conference on computer vision and
  pattern recognition}, pages 770--778, 2016.

\bibitem{hinton2015distilling}
Geoffrey Hinton, Oriol Vinyals, Jeff Dean, et~al.
\newblock Distilling the knowledge in a neural network.
\newblock {\em arXiv preprint arXiv:1503.02531}, 2(7), 2015.

\bibitem{hochba1997approximation}
Dorit~S Hochba.
\newblock Approximation algorithms for np-hard problems.
\newblock {\em ACM Sigact News}, 28(2):40--52, 1997.

\bibitem{hou2021coordinate}
Qibin Hou, Daquan Zhou, and Jiashi Feng.
\newblock Coordinate attention for efficient mobile network design.
\newblock In {\em Proceedings of the IEEE/CVF conference on computer vision and
  pattern recognition}, pages 13713--13722, 2021.

\bibitem{howard2019searching}
Andrew Howard, Mark Sandler, Grace Chu, Liang-Chieh Chen, Bo~Chen, Mingxing
  Tan, Weijun Wang, Yukun Zhu, Ruoming Pang, Vijay Vasudevan, et~al.
\newblock Searching for mobilenetv3.
\newblock In {\em Proceedings of the IEEE/CVF International Conference on
  Computer Vision}, pages 1314--1324, 2019.

\bibitem{howard2018universal}
Jeremy Howard and Sebastian Ruder.
\newblock Universal language model fine-tuning for text classification.
\newblock {\em arXiv preprint arXiv:1801.06146}, 2018.

\bibitem{karypis2000multilevel}
George Karypis and Vipin Kumar.
\newblock Multilevel k-way hypergraph partitioning.
\newblock {\em VLSI design}, 11(3):285--300, 2000.

\bibitem{kendall1938new}
Maurice~G Kendall.
\newblock A new measure of rank correlation.
\newblock {\em Biometrika}, 30(1/2):81--93, 1938.

\bibitem{kernighan1970efficient}
Brian~W Kernighan and Shen Lin.
\newblock An efficient heuristic procedure for partitioning graphs.
\newblock {\em The Bell system technical journal}, 49(2):291--307, 1970.

\bibitem{kolesnikov2020big}
Alexander Kolesnikov, Lucas Beyer, Xiaohua Zhai, Joan Puigcerver, Jessica Yung,
  Sylvain Gelly, and Neil Houlsby.
\newblock Big transfer (bit): General visual representation learning.
\newblock In {\em European conference on computer vision}, pages 491--507.
  Springer, 2020.

\bibitem{kornblith2019similarity}
Simon Kornblith, Mohammad Norouzi, Honglak Lee, and Geoffrey Hinton.
\newblock Similarity of neural network representations revisited.
\newblock In {\em International Conference on Machine Learning}, pages
  3519--3529. PMLR, 2019.

\bibitem{kornblith2019better}
Simon Kornblith, Jonathon Shlens, and Quoc~V Le.
\newblock Do better imagenet models transfer better?
\newblock In {\em Proceedings of the IEEE/CVF conference on computer vision and
  pattern recognition}, pages 2661--2671, 2019.

\bibitem{KrauseStarkDengFei-Fei_3DRR2013}
Jonathan Krause, Michael Stark, Jia Deng, and Li~Fei-Fei.
\newblock 3d object representations for fine-grained categorization.
\newblock In {\em 4th International IEEE Workshop on 3D Representation and
  Recognition (3dRR-13)}, Sydney, Australia, 2013.

\bibitem{krizhevsky2009learning}
Alex Krizhevsky, Geoffrey Hinton, et~al.
\newblock Learning multiple layers of features from tiny images.
\newblock 2009.

\bibitem{lenc2015understanding}
Karel Lenc and Andrea Vedaldi.
\newblock Understanding image representations by measuring their equivariance
  and equivalence.
\newblock In {\em Proceedings of the IEEE conference on computer vision and
  pattern recognition}, pages 991--999, 2015.

\bibitem{li2019delta}
Xingjian Li, Haoyi Xiong, Hanchao Wang, Yuxuan Rao, Liping Liu, Zeyu Chen, and
  Jun Huan.
\newblock Delta: Deep learning transfer using feature map with attention for
  convolutional networks.
\newblock {\em arXiv preprint arXiv:1901.09229}, 2019.

\bibitem{LiuHuihui21}
Huihui Liu, Yiding Yang, and Xinchao Wang.
\newblock Overcoming catastrophic forgetting in graph neural networks.
\newblock In {\em Proceedings of the AAAI conference on artificial
  intelligence}, 2021.

\bibitem{liu2021Swin}
Ze~Liu, Yutong Lin, Yue Cao, Han Hu, Yixuan Wei, Zheng Zhang, Stephen Lin, and
  Baining Guo.
\newblock Swin transformer: Hierarchical vision transformer using shifted
  windows.
\newblock {\em International Conference on Computer Vision (ICCV)}, 2021.

\bibitem{loshchilov2017decoupled}
Ilya Loshchilov and Frank Hutter.
\newblock Decoupled weight decay regularization.
\newblock {\em arXiv preprint arXiv:1711.05101}, 2017.

\bibitem{macqueen1967some}
James MacQueen et~al.
\newblock Some methods for classification and analysis of multivariate
  observations.
\newblock In {\em Proceedings of the fifth Berkeley symposium on mathematical
  statistics and probability}, volume~1, pages 281--297. Oakland, CA, USA,
  1967.

\bibitem{maji13fine-grained}
S.~Maji, J.~Kannala, E.~Rahtu, M.~Blaschko, and A.~Vedaldi.
\newblock Fine-grained visual classification of aircraft.
\newblock Technical report, 2013.

\bibitem{mehta2021mobilevit}
Sachin Mehta and Mohammad Rastegari.
\newblock Mobilevit: light-weight, general-purpose, and mobile-friendly vision
  transformer.
\newblock {\em arXiv preprint arXiv:2110.02178}, 2021.

\bibitem{mellor2021neural}
Joe Mellor, Jack Turner, Amos Storkey, and Elliot~J Crowley.
\newblock Neural architecture search without training.
\newblock In {\em International Conference on Machine Learning}, pages
  7588--7598. PMLR, 2021.

\bibitem{montufar2014number}
Guido~F Montufar, Razvan Pascanu, Kyunghyun Cho, and Yoshua Bengio.
\newblock On the number of linear regions of deep neural networks.
\newblock {\em Advances in neural information processing systems}, 27, 2014.

\bibitem{nguyen2020leep}
Cuong Nguyen, Tal Hassner, Matthias Seeger, and Cedric Archambeau.
\newblock Leep: A new measure to evaluate transferability of learned
  representations.
\newblock In {\em International Conference on Machine Learning}, pages
  7294--7305. PMLR, 2020.

\bibitem{nguyen2021model}
Dang Nguyen, Khai Nguyen, Dinh Phung, Hung Bui, and Nhat Ho.
\newblock Model fusion of heterogeneous neural networks via cross-layer
  alignment.
\newblock {\em arXiv preprint arXiv:2110.15538}, 2021.

\bibitem{nilsback2008automated}
Maria-Elena Nilsback and Andrew Zisserman.
\newblock Automated flower classification over a large number of classes.
\newblock In {\em 2008 Sixth Indian Conference on Computer Vision, Graphics \&
  Image Processing}, pages 722--729. IEEE, 2008.

\bibitem{papadimitriou1998combinatorial}
Christos~H Papadimitriou and Kenneth Steiglitz.
\newblock {\em Combinatorial optimization: algorithms and complexity}.
\newblock Courier Corporation, 1998.

\bibitem{parkhi12a}
O.~M. Parkhi, A.~Vedaldi, A.~Zisserman, and C.~V. Jawahar.
\newblock Cats and dogs.
\newblock In {\em IEEE Conference on Computer Vision and Pattern Recognition},
  2012.

\bibitem{radosavovic2020designing}
Ilija Radosavovic, Raj~Prateek Kosaraju, Ross Girshick, Kaiming He, and Piotr
  Doll{\'a}r.
\newblock Designing network design spaces.
\newblock In {\em Proceedings of the IEEE/CVF Conference on Computer Vision and
  Pattern Recognition}, pages 10428--10436, 2020.

\bibitem{raghu2017svcca}
Maithra Raghu, Justin Gilmer, Jason Yosinski, and Jascha Sohl-Dickstein.
\newblock Svcca: Singular vector canonical correlation analysis for deep
  learning dynamics and interpretability.
\newblock {\em Advances in neural information processing systems}, 30, 2017.

\bibitem{ramsay1984matrix}
JO~Ramsay, Jos ten Berge, and GPH Styan.
\newblock Matrix correlation.
\newblock {\em Psychometrika}, 49(3):403--423, 1984.

\bibitem{ridnik2021imagenet21k}
Tal Ridnik, Emanuel Ben-Baruch, Asaf Noy, and Lihi Zelnik-Manor.
\newblock Imagenet-21k pretraining for the masses, 2021.

\bibitem{russakovsky2015imagenet}
Olga Russakovsky, Jia Deng, Hao Su, Jonathan Krause, Sanjeev Satheesh, Sean Ma,
  Zhiheng Huang, Andrej Karpathy, Aditya Khosla, Michael Bernstein, et~al.
\newblock Imagenet large scale visual recognition challenge.
\newblock {\em International journal of computer vision}, 115(3):211--252,
  2015.

\bibitem{sanh2019distilbert}
Victor Sanh, Lysandre Debut, Julien Chaumond, and Thomas Wolf.
\newblock Distilbert, a distilled version of bert: smaller, faster, cheaper and
  lighter.
\newblock {\em arXiv preprint arXiv:1910.01108}, 2019.

\bibitem{shu2021zoo}
Yang Shu, Zhi Kou, Zhangjie Cao, Jianmin Wang, and Mingsheng Long.
\newblock Zoo-tuning: Adaptive transfer from a zoo of models.
\newblock In {\em International Conference on Machine Learning}, pages
  9626--9637. PMLR, 2021.

\bibitem{singh2020model}
Sidak~Pal Singh and Martin Jaggi.
\newblock Model fusion via optimal transport.
\newblock {\em Advances in Neural Information Processing Systems},
  33:22045--22055, 2020.

\bibitem{SongJieNuerIPS19}
Jie Song, Yixin Chen, Xinchao Wang, Chengchao Shen, and Mingli Song.
\newblock Deep model transferability from attribution maps.
\newblock In {\em Advances in Neural Information Processing Systems}, 2019.

\bibitem{steiner2021train}
Andreas Steiner, Alexander Kolesnikov, Xiaohua Zhai, Ross Wightman, Jakob
  Uszkoreit, and Lucas Beyer.
\newblock How to train your vit? data, augmentation, and regularization in
  vision transformers.
\newblock {\em arXiv preprint arXiv:2106.10270}, 2021.

\bibitem{tran2019transferability}
Anh~T Tran, Cuong~V Nguyen, and Tal Hassner.
\newblock Transferability and hardness of supervised classification tasks.
\newblock In {\em Proceedings of the IEEE/CVF International Conference on
  Computer Vision}, pages 1395--1405, 2019.

\bibitem{van2021benchmarking}
Grant Van~Horn, Elijah Cole, Sara Beery, Kimberly Wilber, Serge Belongie, and
  Oisin Mac~Aodha.
\newblock Benchmarking representation learning for natural world image
  collections.
\newblock In {\em Proceedings of the IEEE/CVF Conference on Computer Vision and
  Pattern Recognition}, pages 12884--12893, 2021.

\bibitem{wang2020federated}
Hongyi Wang, Mikhail Yurochkin, Yuekai Sun, Dimitris Papailiopoulos, and
  Yasaman Khazaeni.
\newblock Federated learning with matched averaging.
\newblock {\em arXiv preprint arXiv:2002.06440}, 2020.

\bibitem{cubbird}
Peter Welinder, Steve Branson, Takeshi Mita, Catherine Wah, Florian Schroff,
  Serge Belongie, and Pietro Perona.
\newblock Caltech-ucsd birds 200.
\newblock Technical Report CNS-TR-201, Caltech, 2010.

\bibitem{williams2021generalized}
Alex Williams, Erin Kunz, Simon Kornblith, and Scott Linderman.
\newblock Generalized shape metrics on neural representations.
\newblock {\em Advances in Neural Information Processing Systems}, 34, 2021.

\bibitem{xiao2021early}
Tete Xiao, Mannat Singh, Eric Mintun, Trevor Darrell, Piotr Doll{\'a}r, and
  Ross Girshick.
\newblock Early convolutions help transformers see better.
\newblock {\em Advances in Neural Information Processing Systems},
  34:30392--30400, 2021.

\bibitem{xie2017aggregated}
Saining Xie, Ross Girshick, Piotr Doll{\'a}r, Zhuowen Tu, and Kaiming He.
\newblock Aggregated residual transformations for deep neural networks.
\newblock In {\em Proceedings of the IEEE conference on computer vision and
  pattern recognition}, pages 1492--1500, 2017.

\bibitem{xuhong2018explicit}
LI~Xuhong, Yves Grandvalet, and Franck Davoine.
\newblock Explicit inductive bias for transfer learning with convolutional
  networks.
\newblock In {\em International Conference on Machine Learning}, pages
  2825--2834. PMLR, 2018.

\bibitem{yamins2014performance}
Daniel~LK Yamins, Ha~Hong, Charles~F Cadieu, Ethan~A Solomon, Darren Seibert,
  and James~J DiCarlo.
\newblock Performance-optimized hierarchical models predict neural responses in
  higher visual cortex.
\newblock {\em Proceedings of the national academy of sciences},
  111(23):8619--8624, 2014.

\bibitem{yang2020transfer}
Xingyi Yang, Xuehai He, Yuxiao Liang, Yue Yang, Shanghang Zhang, and Pengtao
  Xie.
\newblock Transfer learning or self-supervised learning? a tale of two
  pretraining paradigms.
\newblock {\em arXiv preprint arXiv:2007.04234}, 2020.

\bibitem{yang2022factorizing}
Xingyi Yang, Jingwen Ye, and Xinchao Wang.
\newblock Factorizing knowledge in neural networks.
\newblock {\em European Conference on Computer Vision}, 2022.

\bibitem{yang2020factorizable}
Yiding Yang, Zunlei Feng, Mingli Song, and Xinchao Wang.
\newblock Factorizable graph convolutional networks.
\newblock {\em Advances in Neural Information Processing Systems},
  33:20286--20296, 2020.

\bibitem{Yang_2020_CVPR}
Yiding Yang, Jiayan Qiu, Mingli Song, Dacheng Tao, and Xinchao Wang.
\newblock Distilling knowledge from graph convolutional networks.
\newblock In {\em Proceedings of the IEEE/CVF Conference on Computer Vision and
  Pattern Recognition (CVPR)}, June 2020.

\bibitem{ye2019student}
Jingwen Ye, Yixin Ji, Xinchao Wang, Kairi Ou, Dapeng Tao, and Mingli Song.
\newblock Student becoming the master: Knowledge amalgamation for joint scene
  parsing, depth estimation, and more.
\newblock In {\em Proceedings of the IEEE/CVF Conference on Computer Vision and
  Pattern Recognition}, pages 2829--2838, 2019.

\bibitem{you2021logme}
Kaichao You, Yong Liu, Jianmin Wang, and Mingsheng Long.
\newblock Logme: Practical assessment of pre-trained models for transfer
  learning.
\newblock In {\em International Conference on Machine Learning}, pages
  12133--12143. PMLR, 2021.

\bibitem{yun2019cutmix}
Sangdoo Yun, Dongyoon Han, Seong~Joon Oh, Sanghyuk Chun, Junsuk Choe, and
  Youngjoon Yoo.
\newblock Cutmix: Regularization strategy to train strong classifiers with
  localizable features.
\newblock In {\em Proceedings of the IEEE/CVF international conference on
  computer vision}, pages 6023--6032, 2019.

\bibitem{zhang2021quantifying}
Guojun Zhang, Han Zhao, Yaoliang Yu, and Pascal Poupart.
\newblock Quantifying and improving transferability in domain generalization.
\newblock {\em Advances in Neural Information Processing Systems}, 34, 2021.

\bibitem{zhang2017mixup}
Hongyi Zhang, Moustapha Cisse, Yann~N Dauphin, and David Lopez-Paz.
\newblock mixup: Beyond empirical risk minimization.
\newblock {\em arXiv preprint arXiv:1710.09412}, 2017.

\bibitem{zhou2020rethinking}
Daquan Zhou, Qibin Hou, Yunpeng Chen, Jiashi Feng, and Shuicheng Yan.
\newblock Rethinking bottleneck structure for efficient mobile network design.
\newblock In {\em European Conference on Computer Vision}, pages 680--697.
  Springer, 2020.

\bibitem{zhou2021autospace}
Daquan Zhou, Xiaojie Jin, Xiaochen Lian, Linjie Yang, Yujing Xue, Qibin Hou,
  and Jiashi Feng.
\newblock Autospace: Neural architecture search with less human interference.
\newblock In {\em Proceedings of the IEEE/CVF International Conference on
  Computer Vision}, pages 337--346, 2021.

\bibitem{zhou2021deepvit}
Daquan Zhou, Bingyi Kang, Xiaojie Jin, Linjie Yang, Xiaochen Lian, Zihang
  Jiang, Qibin Hou, and Jiashi Feng.
\newblock Deepvit: Towards deeper vision transformer.
\newblock {\em arXiv preprint arXiv:2103.11886}, 2021.

\bibitem{zhou2021ensemble}
Zhi-Hua Zhou.
\newblock Ensemble learning.
\newblock In {\em Machine learning}, pages 181--210. Springer, 2021.

\end{thebibliography}




\end{document}


\maketitle

In this document, we provide additional materials
that cannot fit into the main manuscript due to the page limit.
We first give the full deviation of the function similarity and the pseudo code for the \name~framework, 
and then conduct additional experiments 
to validate the proposed \name. 
Next, we describe our implementation details, 
dataset settings, evaluation metrics, and hyper-parameter settings.

\section{Pseudo-code for \name}
Algorithm~\ref{alg:pseudo} provides the pseudo-code of our two-stage solution for \name. Given a model zoo $Z$, we run the joint network partition for $R$ times. Each partition stage involves a tri-level optimization, with a block partition step, an anchor selection step, and a block assignment step. The optimal network partition 
yields a number of build blocks, which
are then taken to reassemble a new network by 
solving a constraint integer program with a training-free proxy.
\begin{algorithm}[H]
\scriptsize
	\caption{Deep Model Reassembly} 
		\label{alg:pseudo}
	\begin{algorithmic}[1]
	\State \textcolor{SeaGreen}{\# Network Partition by Functional Equivalence}
		\For {$run=1,2,\ldots,R$}
    		\State Initialize the network partition $\{B_{i}^{(k)}\}_{k=1}^K|_{0}$ and clustering $A|_0$
			\For {$t=1,2,\ldots, T$  and $J(A|_{t}, \{B_{i}^{(k)}\}_{k=1}^K|_{t}) - J(A|_{t-1}, \{B_{i}^{(k)}\}_{k=1}^K|_{t-1}) \leq \epsilon$}
				\State Update the partition by network layer swapping $\{B_{i}^{(k)}\}_{k=1}^K|_{t} \leftarrow \arg \max_{B_{i}^{(k)}} J(A|_{t-1}, \{B_{i}^{(k)}\}_{k=1}^K|_{t-1})$ 
				\State Update the anchor blocks $B_{a_j}|_{t} \leftarrow \arg \max_{B_{a_j}} J(A|_{t-1}, \{B_{i}^{(k)}\}_{k=1}^K|_{t})$ 
				\State Update the block assignment $A|_{t} \leftarrow \arg \max_{A} J(A|_{t-1}, \{B_{i}^{(k)}\}_{k=1}^K|_{t})$ 
			\EndFor
		\EndFor
		\State Find the best partition in all runs $A^*,\{B^*_{i}^{(k)}\}_{k=1}^K \leftarrow \arg \max_{A, B_{i}^{(k)}} J(A, \{B_{i}^{(k)}\}_{k=1}^K)$.
		\State \textcolor{SeaGreen}{\# Network Reassembly by Solving an Integer Program}
		\For {$c=1,2,\ldots,num\_candidate$}
		    \State Reassemble a network randomly $M(X,Y)$ under hard constraints.
		    \State Compute the training-free score  \texttt{Scores}[c] $\leftarrow$ NASWOT($M(X,Y)$)
		\EndFor  
		\State Find the best candidate with maximum score $c^*\leftarrow\arg \max_{c}$ \texttt{Scores}
	\end{algorithmic} 

\end{algorithm}

\section{Functional similarity}
\subsection{Full Derivation of Function Similarity}
Following  Def.~3, we are able to 
find the best proxy $B'$ of a given neural network $B$ by solving
{\footnotesize
\begin{align}
    B'^* = \arg\max_{B'} s(B(\mathbf{X}), B'(\mathbf{X}')), \quad s.t.\quad s(\mathbf{X}, \mathbf{X}') - \epsilon \geq 0
\end{align}}\noindent
Using Lagrange multipliers, our objective can be simplified to $L(B', \lambda) = s(B(\mathbf{X}), B'(\mathbf{X}'))+\lambda s(\mathbf{X}, \mathbf{X}') - \lambda\epsilon$. By omitting the last term~(not related to our $B'$) and setting $\lambda=1$, we have $S(B,B') = s(B(\mathbf{X}), B'(\mathbf{X}'))+ s(\mathbf{X}, \mathbf{X}')$ to characterize the function similarity of two neural blocks, which is a weighted summation of its input-output similarity. 

\subsection{Function Similarity and Knowledge Distillation}
Knowledge distillation~(KD)~\cite{hinton2015distilling} aims to learn a compact student network $S:\mathbb{R}^{d_{in}} \to \mathbb{R}^{d_{out}}$ from the teacher network $T:\mathbb{R}^{d'_{in}} \to \mathbb{R}^{d'_{out}}$.
For the typical KD problem where $d_{in}=d'_{in}$ and $d_{out}=d'_{out}$, given the same input, we would like to minimize the difference between two models' output logits. Specifically, 
\begin{itemize}[itemsep=2pt,parsep=2pt,leftmargin=1em,labelwidth=1em,labelsep=2pt]
    \item If we adopt \textbf{mean-square-error}~(MSE), it
     measures the functional similarity using linear regression $R^2_{\text{LR}}$~\cite{kornblith2019similarity}. 
    \item If we adopt  \textbf{KL divergence} $\mathbb{D}_{KL}\big(S(\mathbf{X})| T(\mathbf{X})\big)$ as our distillation loss, it matches well with the functional similarity with mutual information $MI\big(S(\mathbf{X}), T(\mathbf{X})\big) = H\big(T(\mathbf{X}) \big)- \mathbb{D}_{KL}\big(S(\mathbf{X})| T(\mathbf{X})\big)$. We can drop the first term since the teacher output entropy $H\big(T(\mathbf{X}) \big)$ is considered as a constant.
\end{itemize}

The analogy can be brought up when distilling intermediate feature~\cite{romero2014fitnets} from teacher to student. Here we assume $d_{in}=d'_{in}$ and $d_{out}\neq d'_{out}$. One typical solution is to add one linear layer $W \in \mathbb{R}^{d_{out} \times d'_{out}}$ on top of the student network to match  the output dimensions and then minimize the $MSE\big(T(\mathbf{X}), S(\mathbf{X})W \big)$. Because the $R^2_{\text{LR}}$ is invariant to invert-able linear transformation~\cite{kornblith2019similarity}, as long as $W$ is invert-able, $s\big(S(\mathbf{X})W, T(\mathbf{X})\big) = s\big(S(\mathbf{X}), T(\mathbf{X})\big)$. It again lies in our definition of functional similarity. If we consider the feature distillation by maximizing the mutual information~\cite{tian2019contrastive, ahn2019variational}, the functional similarity with mutual information describes any form of non-linear transformation instead of a single linear layer. 

In sum,  KD can be considered as a special case for our functional similarly, when $d_{in}=d'_{in}$.

\subsection{Function Similarity of Two Identical Networks}
We now prove that function similarity is the necessary but insufficient condition for two identical neural networks. By the term \emph{identical}, we mean the two networks have exactly the same architecture as well as the weights. Let us assume we have an arbitrary representation similarity $s(\cdot, \cdot) \in [0,1]$ and its corresponding functional similarly $S(\cdot, \cdot) \in [0,2]$, it is clear that $s(\mathbf{X}, \mathbf{X})=1$. 
 We show that for two networks $B$ and $B'$, (1) [$B,B'\text{ are identical} \Rightarrow S(B, B')=2$] but (2) [$S(B, B')=2  \not\Rightarrow B,B'\text{ are identical}$].

\begin{itemize}[itemsep=2pt,parsep=2pt,leftmargin=1em,labelwidth=1em,labelsep=2pt]
    \item \textbf{Necessity.} If two networks are identical, given the same input $\mathbf{X}$, two networks ought to produce the same output $B(\mathbf{X}) = B'(\mathbf{X})$, with $s(B(\mathbf{X}), B'(\mathbf{X}))=1$. Then, we have $S(B,B') = s(B(\mathbf{X}), B'(\mathbf{X})) + s(\mathbf{X}, \mathbf{X})=2$.
    \item \textbf{Insufficiency.} We prove that [$S(B, B')=2  \not\Rightarrow B,B'\text{ are identical}$] by contradiction. Suppose $B$ is a single-layer linear neural network $B(\mathbf{X}) = W\mathbf{X}$, and $B'$ is 3-layer network with a scaling and a descaling layer $B'(\mathbf{X}) = (kI) W (\frac{1}{k} I) \mathbf{X}$. Though the two networks have the same output given the same input, they are indeed not identical.
\end{itemize}

\typo{\subsection{Computational Complexity Analysis and Solutions}}
\typo{In our formulation, we need to compute the feature similarity within each equivalence set. If we do it along side with the network partition, each network need to be forwarded to get the intermediate feature  at every iteration, and then we compute the similarity between $K \times N$ blocks and $K$ centroids for three times (one for clustering and two for a forward and a backward layer swapping). Assume there is $n$ data points~($\frac{1}{20}$ of the full dataset). We run the optimization in Section 3.2 for $t$ steps for each run and repeat the experiment for $R$ times, the time complexity for representation similarity is $\mathcal{O}(K \times N \times K \times t \times R)$ and the time complexity for network inference is $\mathcal{O}(N \times n \times R)$. At the same time, similarity computation for each pair of block is huge. Taking linear CKA as an example, given two feature matrix $X \in \mathbb{R}^{n \times d_1}$ and $Y \in \mathbb{R}^{n \times d_2}$, the computation of linear CKA envolves 3 matrix multiplication}
\typo{\begin{align}
    CKA(XX^\top, YY^\top) & = \frac{||Y^\top X||_F^2}{||XX^\top||_F ||YY^\top||_F } 
\end{align}}

\typo{Since the number of samples and feature dimension is large~($n>10000, d>1000$), the memory consumption is extremely large. All above factors inevitably leads to huge computational over head. }

\typo{As the \textbf{online computation} is cumbersome in practice, we take an alternative path to fill in the feature similarity table \textbf{offline}. We take 3 strategies to significantly reduces the computational complexity in our implementation.}
\begin{enumerate}
    \item \typo{\textbf{Reduce the Inference Time.} We perform data inference on $n$ samples offline, and save the intermediate feature vectors to local files. This reduce the network inference complexity from $\mathcal{O}(N \times n \times R)$ to $\mathcal{O}(N \times n)$.}
    \item \typo{\textbf{Reduce the Similarity Computation.} We compute the similarity for all networks for $\sum_{i=1}^N\sum_{j=1}^N L_i \times L_j$ times for each pair of layers. This reduces the time complexity from $\mathcal{O}(K \times N \times K \times t \times R)$ to $\mathcal{O}(\sum_{i=1}^N\sum_{j=1}^N L_i \times L_j)$. This seems to be trivial, however, when we assume  $L=20$, $t=100$, $R=200$, $K=4$, $N=28$, the online version needs $4\times 4\times 28 \times 100 \times 200 = 8,960,000$ computations, and the offline version gets $28 \times 28 \times 20 \times 20=313,600$ times. In fact, for each block similarity in online computation, we need to do the pair-wise computation for three times~(Line 58). There is a $1\sim 2$ orders of magnitude speed up.}
    \item \typo{\textbf{Reduce the Memory for Large Matrix Multiplication.} Since there involves large matrix multiplication, the memory requirement is extremely large. We implement a mini-batch version full CKA algorithm to save memory and do multi-thresholding. In~\cite{nguyen2020wide}, it has been proved that the mini-batch CKA provides unbiased estimator of CKA value. We also implement a torch-version CKA on GPU to accelerate the computation.}
\end{enumerate}
\typo{Then, the offline computation for each pair of networks reduces to around $1 \sim 2$ min. For a zoo of 28 models, we need around $28\times 28 \times 1 \sim 28\times 28 \times 2$ minutes, with about 12-24 hour in total. Once the similarity computation is done, the partition and reassembly steps can be done without hassle. Please review our code at \texttt{similarity/get\_rep.py} and \texttt{similarity/compute\_sim.py} on the implementation details.}

\typon{\section{Limitation, Social Impact and Possible Solutions}}
\typon{Currently, our \name has several limitations, especially with regards to \emph{privacy concerns} and \emph{model bias}. The following discussion of each of those factors provides a more comprehensive reflection on the current research.}

\begin{itemize}
    \item \typon{\textbf{Privacy Concern.} \name~combines multiple trained models by reusing some of their parameters. Thus, the reassembled model could contain information from multiple models. Criminals may embezzle the private information or intellectual property of the upstream model by attacking the reassembled model. The problem might be solved by increasing the number of trained models so that the reserved parameters of each model are not sufficient to recover the complete model information. }  
    \item \typon{\textbf{Model Bias.} While \name~inherits the privileged knowledge of its predecessors, the data bias and knowledge bias may also be transferred to the reassembled model. To tackle this problem, two techniques could be incorporated into the \name~framework to mitigate the predecessor model bias. \textbf{For starters}, we can expand the model zoo size and limit the assembled size for each model. It ensures that no single model bias is dominant in the reassembled model. \textbf{It is also possible} to increase the diversity among the reassembled blocks instead of blindly optimizing the target performance. A diversity-promoting regularization term could be added to Equation 8 to provide unbiased predictions. }  
\end{itemize}
\typon{We will extend our study to those fields to eliminate the current limitation of \name. }
\typon{\section{Potential Applications}}
\typon{As \name~provides a framework to integrate the trained networks and produce a new model, we discuss potential applications on \emph{large model training} and \emph{multi-task learning}.}
\begin{itemize}
    \item \typon{\textbf{Large model training.} Large models have recently shown their remarkable potential in building powerful AI systems with unprecedented generalization ability and robustness capability. However, training a large model is extremely cumbersome and computationally intensive. Instead of training a network from scratch, \name~allows us to take several pre-trained small models, partition them into building blocks and assemble them into the large model as an efficient method for network initialization. As suggested in the paper, reassembling pre-trained models provides faster convergence and reduces the training cost.}
    \item \typon{\textbf{Multi-task Learning.} Typical multi-tasking learning requires training a model on a complex task combination. \name~shed light on a very cheap alternative method for multi-task training. Given a bunch of trained single-task models, we can develop a method to aggregate their capacities into a reassembled model. For example, assemble a new model with a shared backbone and multiple task prediction components, each taken from a single task. As such, we reassemble a multi-task model at a very low cost using \name.} 
\end{itemize}

\section{Additional Experimental Results}
\typon{\subsection{Heterogeneous Models VS Homogeneous Models}}
\typon{Although our \name~was originally designed for fusing heterogeneous models, we would like extend it to homogeneous models to demonstrate its generality.  We, therefore, adopt \name~on homogeneous models and compare the results with those of the Zoo Tuning~\cite{shu2021zoo} on CIFAR-100, AirCraft and Cars using the same homogeneous zoo setting as in~\cite{shu2021zoo}. Note that we do not further pre-train \name~on ImageNet to make sure the comparison is fair. As shown in Table~\ref{tab:homogeneous model zoo}, We improve the accuracy by 4.13\% on Cars and 3.35\% on AirCraft dataset, with marginal computational overhead. We indeed outperform~\cite{shu2021zoo} significantly with the same experimental setup. }

\begin{table}[H]
    \centering
    \begin{tabular}{l|l|l|l|l}
    \toprule
         & # Param(M) & CIFAR-100 & AirCraft & Cars\\
         \hline
        Zoo Tuning & 23.71 &	83.39&	85.51&	89.73\\
        \name(4, 30, 6) & 24.89	&\textbf{84.05}(\textcolor{Green}{+0.66})&	\textbf{88.86}(\textcolor{Green}{+3.35})& \textbf{93.86}(\textcolor{Green}{+4.13})\\
        \bottomrule
    \end{tabular}
    \caption{\typon{Performance and computation comparison on homogeneous model zoo.}}
    \label{tab:homogeneous model zoo}
    \vspace{-5mm}
\end{table}
\typon{In fact, a homogeneous model zoo is a simplified case for heterogeneous models. In our main paper, we have focused on the more challenging heterogeneous models.}
\typon{\subsection{Ablation Study on Partition Number, Performance and Complexity}}
\typon{In our main paper, we assume the network partition number $K=4$. We repeat the network partition and reassembly steps with $K=5$ and $K=6$ to see how our method performs with different partition granularities. We set the configuration to \name(K, 30, 6). The network is trained on ImageNet for 100 epochs. All other experimental settings remain the same as those used in the main paper. As illustrated in Table~\ref{tab:Ablation_partition}, we observe that the partition number $K$ has a minimal effect on the model performance. As for the computational complexity, a large $K$ does increase the search time with less than 10\% time growth. This ablation study suggests that our proposed \name~is not sensitive to the selection of partition number $K$.}

\begin{table}[H]
    \centering
    \begin{tabular}{c|l|l|l|l}
    \toprule
         K & #Param(M) & GFLOPs & Top1 Acc & Top 5 Acc  \\
    \hline
         4&	24.89 & 4.47 & 79.63 & 94.81 \\
         5&	21.14 & 5.53 & 79.68 & 94.89 \\
         6&	23.38 & 5.39 & 79.83 & 95.02 \\
         \bottomrule
    \end{tabular}
    \caption{\typon{Ablation study for the partition number $K$.}}
    \label{tab:Ablation_partition}
\end{table}

\subsection{Partition Results}
\begin{figure}[H]
    \centering
   \subfigure[$K=4$]{ \includegraphics[width=0.7\linewidth]{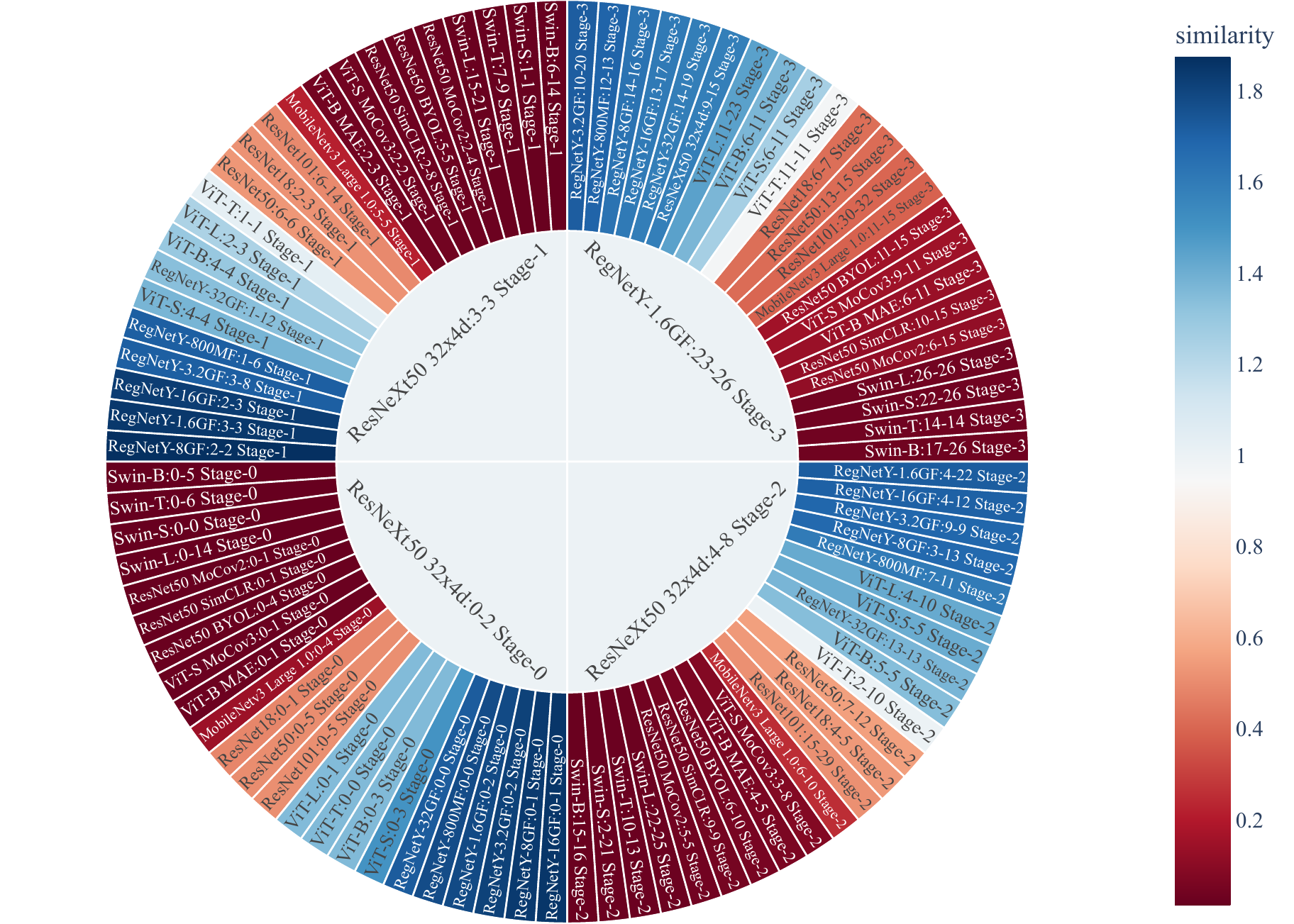}}
   \subfigure[$K=6$]{ \includegraphics[width=0.7\linewidth]{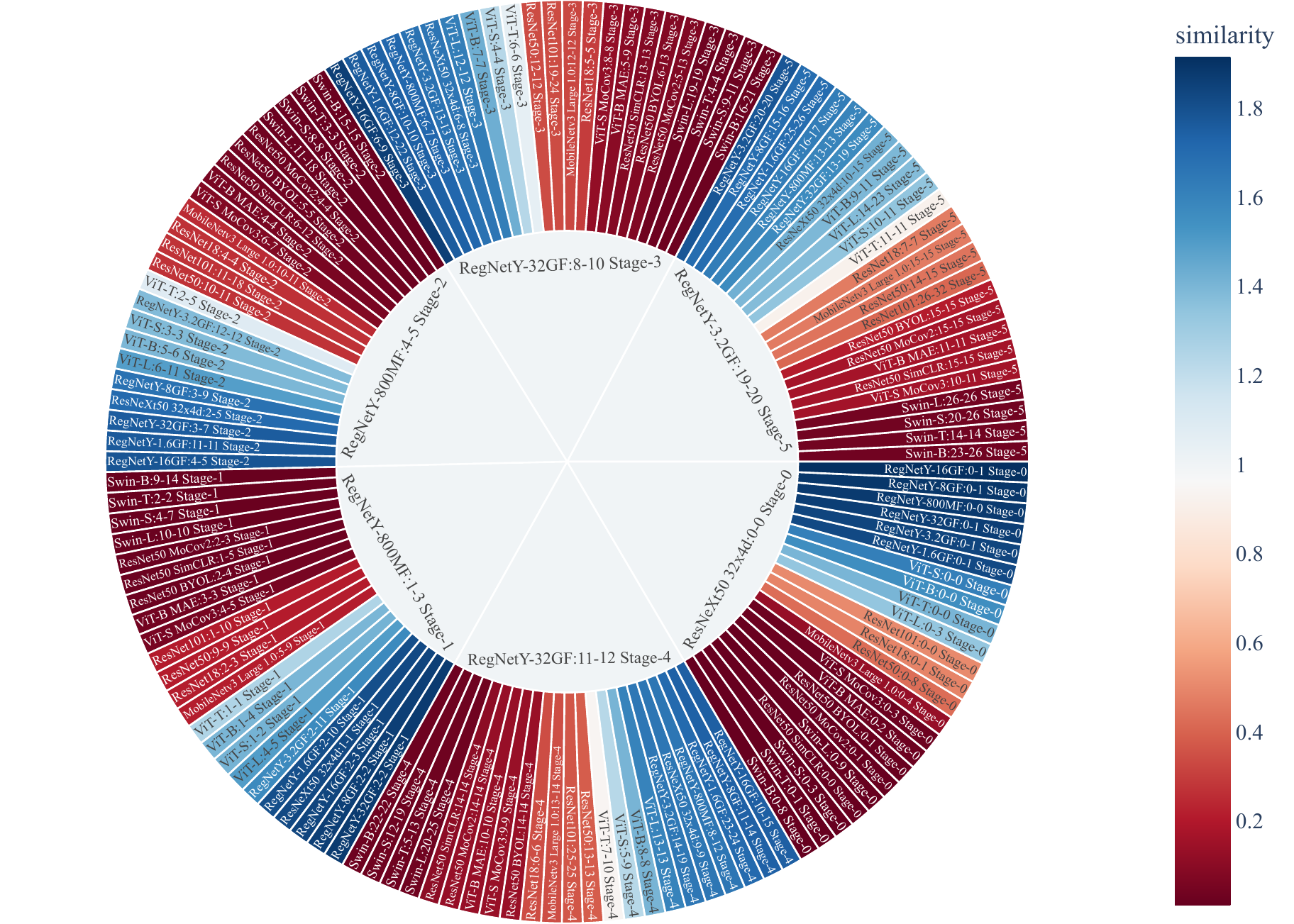}}
    \caption{Model partition and assignment visualization.}
    \label{fig:partition_vis}
\end{figure}

We visualize the network partition hierarchy in Figure~\ref{fig:partition_vis} with linear CKA when the partition number $K=4$ and $K=6$. 
We name each block in the format of
\texttt{ModelName}-\texttt{NodeRange}-\texttt{StageIndex}. The anchor blocks are shown in the inner circle, while the outside ring lists all of the blocks in each equivalence set. The panel color indicates the functional similarity between each block and its anchor block. 

We make the following observations
\begin{enumerate}[itemsep=2pt,parsep=2pt,leftmargin=1em,labelwidth=1em,labelsep=2pt]
    \item The equivalent sets tend to \emph{cluster the blocks by stage index}. For example, all Stage 0 blocks of various pre-trained networks are within the same equivalence set. It provides valuable insight that neural networks learn similar patterns at similar network stages, despite of its architecture or training strategy.
    \item Our network partition prefers ResNeXt- and RegNet-structured network as its anchor blocks.
\end{enumerate}

\subsection{Architecture or Pre-trained Weights?}
\begin{figure}[H]
    \centering
    \includegraphics[width=\linewidth]{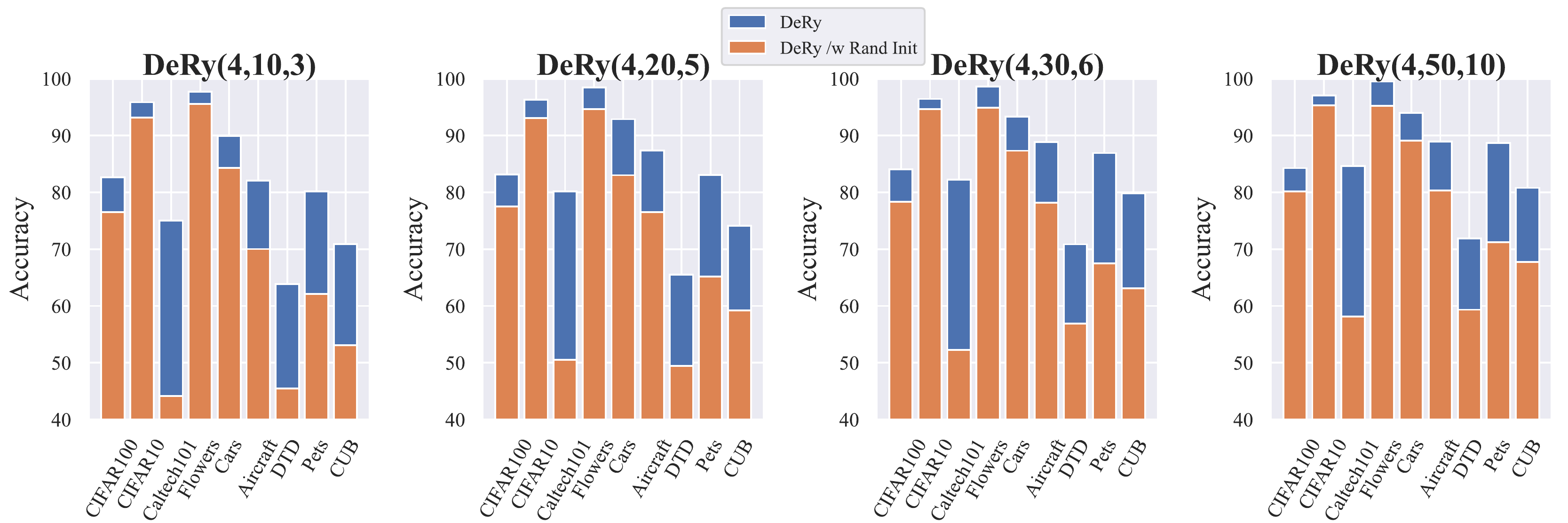}
    \caption{Performance comparison between \name~and \name~with random weight initialization.}
    \label{fig:arch_weight}
\end{figure}

We would like to investigate whether it is the searched architecture alone, 
or \emph{both} the network structure \emph{and} the pre-trained weights 
contribute to the performance improvement in our \name~solution. 
Figure~\ref{fig:arch_weight} compares the test accuracy between full \name~and \name~with random weight initialization on 9 transfer learning datasets. 
We observe that the \name~model trained from scratch 
does not gain satisfying transfer performance. 
It indicates that the searched architecture and the pre-trained 
weights bring about the performance promotion collaboratively.
\subsection{Visualizing the Representation Similarity}
Figure~\ref{fig:feat sim} visualizes the pairwise network representation similarity heatmap. We include ImageNet1k supervised ResNet-18~(R18), ResNet-50~(R50),  ResNet-101~(R101), ResNeXt-50~(RX50), MobileNetv3 Large~(MBNv3), Vision Transformer Large~(ViT-L), ResNetY-8G~(RegY8G), ImageNet1k BYOL ResNet-50~(R50BYOL), and ImageNet1k MoCov3 Vision Transformer Base~(ViT-B MoCov3). We have the following observations
\begin{itemize}[itemsep=2pt,parsep=2pt,leftmargin=1em,labelwidth=1em,labelsep=2pt]
    \item \textbf{Diagonal Pattern.} For models trained with diverse network structures, random seeds, and training recipes, representations at similar depth generally have high similarity, resulting in the diagonal pattern in each pairwise heatmap. It again indicates that neural networks learn similar patterns at similar depths.
    \item \textbf{Self-supervised Models Learn High-level Semantics.} As shown in the last row~(ViT-B MoCov3) and last column~(R50 BYOL) in Figure~\ref{fig:feat sim}, the self-supervised models achieve high representation similarity with the supervised models at the top layers~(top-right on each heatmap). It suggests that the self-supervised learning might be able to learn the high-level semantics of the supervised pre-training. On the contrary, two training paradigms learn distinctive low-level patterns.
\end{itemize}
\begin{figure}
    \centering
     \subfigure[\tiny R50-R50]{\includegraphics[width=0.19\linewidth]{fig/sim_plot/resnet50.resnet50.pdf}
    }
    \subfigure[\tiny R50-R18]{\includegraphics[width=0.19\linewidth]{fig/sim_plot/resnet18.resnet50.pdf}
    }
    \subfigure[\tiny R50-R101]{\includegraphics[width=0.19\linewidth]{fig/sim_plot/resnet50.resnet101.pdf}}
    \subfigure[\tiny R50-ViT-L]{\includegraphics[width=0.19\linewidth]{fig/sim_plot/resnet50.vit_large_patch16_224.pdf}
    }
    \subfigure[\tiny R50-R50BYOL]{\includegraphics[width=0.19\linewidth]{fig/sim_plot/resnet50.resnet50byol.pdf}
    }
        \hline
    \subfigure[\tiny RX50-RX50]{\includegraphics[width=0.188\linewidth]{fig/sim_plot/swsl_resnext50_32x4d.swsl_resnext50_32x4d.pdf}
    }
    \subfigure[\tiny RX50-R18]{\includegraphics[width=0.188\linewidth]{fig/sim_plot/resnet18.swsl_resnext50_32x4d.pdf}
    }
    \subfigure[\tiny RX50-R101]{\includegraphics[width=0.188\linewidth]{fig/sim_plot/swsl_resnext50_32x4d.resnet101.pdf}
    }
    \subfigure[\tiny RX50-ViT-L]{\includegraphics[width=0.188\linewidth]{fig/sim_plot/swsl_resnext50_32x4d.vit_large_patch16_224.pdf}
    }
    \subfigure[\tiny RX50-R50BYOL]{\includegraphics[width=0.188\linewidth]{fig/sim_plot/swsl_resnext50_32x4d.resnet50byol.pdf}
    }
    \hline
    \subfigure[\tiny MBNv3-MBNv3]{\includegraphics[width=0.188\linewidth]{fig/sim_plot/mobilenetv3_large_100.mobilenetv3_large_100.pdf}
    }
        \subfigure[\tiny MBNv3-R18]{\includegraphics[width=0.188\linewidth]{fig/sim_plot/resnet18.mobilenetv3_large_100.pdf}
    }
    \subfigure[\tiny MBNv3-R101]{\includegraphics[width=0.188\linewidth]{fig/sim_plot/mobilenetv3_large_100.resnet101.pdf}
    }
    \subfigure[\tiny{MBNv3-ViT-L}]{\includegraphics[width=0.188\linewidth]{fig/sim_plot/mobilenetv3_large_100.vit_large_patch16_224.pdf}
    }
    \subfigure[\tiny{MBNv3-R50BYOL}]{\includegraphics[width=0.188\linewidth]{fig/sim_plot/mobilenetv3_large_100.resnet50byol.pdf}
    }
        \hline
    \subfigure[\tiny RegY8G-RegY8G]{\includegraphics[width=0.188\linewidth]{fig/sim_plot/regnet_y_8gf.regnet_y_8gf.pdf}
    }
        \subfigure[\tiny{RegY8G-R18}]{\includegraphics[width=0.188\linewidth]{fig/sim_plot/resnet18.regnet_y_8gf.pdf}
    }
    \subfigure[\tiny{RegY8G-R101}]{\includegraphics[width=0.188\linewidth]{fig/sim_plot/regnet_y_8gf.resnet101.pdf}
    }
    \subfigure[\tiny{RegY8G-ViT-L}]{\includegraphics[width=0.188\linewidth]{fig/sim_plot/regnet_y_8gf.vit_large_patch16_224.pdf}
    }
    \subfigure[\tiny{RegY8G-R50BYOL}]{\includegraphics[width=0.188\linewidth]{fig/sim_plot/resnet50byol.regnet_y_8gf.pdf}
    }
            \hline
    \subfigure[\tiny{ViT-B Mv3-ViT-B Mv3} ]{\includegraphics[width=0.188\linewidth]{fig/sim_plot/vit_base_patch16_224mocov3.vit_base_patch16_224mocov3.pdf}
    }
        \subfigure[\tiny{ViT-B MoCov3-R18}]{\includegraphics[width=0.188\linewidth]{fig/sim_plot/resnet18.vit_base_patch16_224mocov3.pdf}
    }
    \subfigure[\tiny{ViT-B MoCov3-R101}]{\includegraphics[width=0.188\linewidth]{fig/sim_plot/vit_base_patch16_224mocov3.resnet101.pdf}
    }
    \subfigure[\tiny{ViT-B MoCov3-ViT-L}]{\includegraphics[width=0.188\linewidth]{fig/sim_plot/vit_base_patch16_224mocov3.vit_large_patch16_224.pdf}
    }
    \subfigure[\tiny{ViT-B MoCov3-R50BYOL}]{\includegraphics[width=0.188\linewidth]{fig/sim_plot/vit_base_patch16_224mocov3.resnet50byol.pdf}
    }
    \caption{Feature similarity heatmap of heterogeneous pre-trained models on ImageNet.}
    \label{fig:feat sim}
\end{figure}
\newpage
\subsection{Numerical Results for Transfer Learning}
Table~\ref{tab:pretrained_table} and Table~\ref{tab:rand_table} provide the numerical performance on 9 tasks from pre-trained weights or from random initialization. The same results are also shown in Figure 11 in the main paper.

\renewcommand{\arraystretch}{1.1}
\setlength{\tabcolsep}{2pt}
\begin{table}[!ht]
\scriptsize
    \centering
    \begin{tabular}{l|l|l|l|l|l|l|l|l|l|l|l}
    \toprule
        Network & Init. & \#\typo{Param}(M) & CIFAR100 & CIFAR10 & Caltech101 & Flowers & Cars & Aircraft & DTD & Pets & CUB  \\ \hline
        
        ResNet-50 & iNat2021 & 23.51 & 84.81 & 97.39 & 92.54 & 97.86 & 91.47 & 84.54 & 72.83 & 91.11 & 82.37  \\ \hline
        ResNet-50 & BYOL & 23.51 & 86.14 & 97.85 & 88.74 & 98.66 & 91.95 & 87.04 & 75.21 & 90.32 & 81.72  \\ \hline
        ResNet-50 & MoCov2 & 23.51 & 83.26 & 96.88 & 80.52 & 98.31 & 92.6 & 88.41 & 68.35 & 86.23 & 79.60  \\ \hline
        ResNet-50 & SimCLR & 23.51 & 82.57 & 96.44 & 85.70 & 98.69 & 91.78 & 83.88 & 71.01 & 88.38 & 85.61  \\ \hline
        ResNet-18 & In1K & 11.69 & 83.00 & 96.17 & 91.70 & 97.26 & 91.00 & 82.70 & 71.60 & 89.90 & 80.75  \\ \hline
        ResNet-50 & In1k & 23.51 & 84.34 & 96.77 & 91.77 & 97.51 & 91.72 & 86.63 & 73.16 & 90.54 & 83.02  \\ \hline
        ResNet-101 & In1K & 42.52 & 86.81 & 97.68 & 93.12 & 97.94 & 91.69 & 85.57 & 74.21 & 92.02 & 83.88  \\
        \hline

        ResNeXt 50 32x4d & In1K & 25.03 & 84.62 & 97.18 & 91.91 & 97.18 & 91.47 & 85.39 & 72.92 & 90.94 & 82.63  \\ \hline
        ResNeXt 101 32x8d & In1K & 88.79 & 85.75 & 97.41 & 93.47 & 97.68 & 91.03 & 84.66 & 72.5 & 90.83 & 82.63 \\ \hline
        MobileNet v3 Large 0.75 & In1K & 4.00 & 80.60 & 95.74 & 89.14 & 96.63 & 91.11 & 82.83 & 70.88 & 88.51 & 79.77  \\ \hline
       MobileNet v3 Large 1.0 & In1K & 5.40 & 82.31 & 96.13 & 91.07 & 97.52 & 91.80 & 86.80 & 71.03 & 88.99 & 81.70  \\ \hline
         RegNetY-800m      & in1k            &  6.30         & 83.58         & 96.57           & 87.16       & 99.47        & 87.90 & 79.99 & 70.96 & 89.19 & 74.82     \\ \hline
        RegNetY-1.6GF      & in1k            & 9.19          & 85.96         & 97.14          & 90.72       & 99.58        & 91.83 & 86.94 & 74.15 & 91.74 & 82.69      \\ \hline
        RegNetY-3.2GF       & in1k            & 20.60         & 85.94          & 94.57          & 86.31         & 98.79          & 89.57   & 83.01   & 74.38   & 87.69   & 84.38       \\ \hline
        RegNetY-8GF       & in1k            &39.20          & 86.23         & 97.72          & 90.91       & 99.31        & 93.04 & 86.38 & 75.37 & 92.57 & 82.93     \\ \hline
        RegNetY-16GF        & in1k            & 83.59         & 87.53          & 96.32          & 89.76         & 98.96          & 92.34   & 85.01   & 75.74   & 87.69   & 84.38       \\\hline
         RegNetY-32GF        & in1k            & 145.05         & 90.92          & 96.87          & 93.31         & 99.69          & 93.51   & 85.01   & 74.38   & 90.69   & 86.38       \\ \hline
          Swin-B  & in21k  & 87.77         & 90.30 & 98.93 & 94.90 & 99.45 & 93.30 & 89.00 & 76.10 & 91.70 & 86.80      \\ \hline
        Swin-T             & in1k            & 28.29         & 86.56         & 97.41          & 92.25         & 98.74          & 91.94   & 83.51   & 67.04   & 86.12    & 85.39       \\ \hline
         Swin-S & in1k            & 49.61            & 89.98         & 98.71           & 93.45       & 99.46        & 91.24 & 85.63 & 78.40 & 93.83 & 86.88      \\ \hline
        Swin-B  & in1k  & 87.77         & 89.52         & 98.41          & 92.76         & 99.04          & 91.94   & 84.51   & 77.82   & 89.14    & 85.97       \\ \hline
        Swin-L  & in1k  & 196.53         & 90.76         & 98.43          & 92.25         & 99.24          & 92.94   & 86.78   & 78.14   & 91.10    & 87.39       \\ \hline
         ViT-S & MoCov3  & 22.88         & 88.29         & 96.16          & 90.62       & 98.97        & 88.63 & 83.76 & 74.33 & 88.15  & 83.11    \\ \hline
        ViT-B              & MAE           & 86.86         & 91.82         & 98.76          & 93.12       & 99.63        & 90.55 & 80.29 & 78.40 & 91.51  & 86.16     \\ \hline
        ViT-T              & In1k            & 5.72          & 87.74         & 98.09          & 89.98       & 99.14        & 83.67 & 69.75 & 73.03 & 88.54 & 80.98     \\ \hline
        ViT-S              & In1k            & 22.88         & 91.82         & 98.76          & 93.13      & 99.63        & 90.55 & 80.29 & 78.40 & 91.51  & 86.16     \\ \hline
        ViT-B              & in1k            &      86.86         & 91.70          & 98.71          & 91.99       & 99.74        & 92.05 & 79.48 & 75.90 & 91.02 & 86.74     \\ \hline
         ViT-L & In1K & 304.72  & 90.65 & 98.34 & 93.21 & 97.35 & 91.98 & 85.33 & 73.42 & 92.47 & 86.73  \\\hline\hline
       \name(4,10,3)-FT & \name+in1k & 7.64 & 84.41 & 96.89 & 88.81 & 99.58 & 93.12 & 86.19 & 73.46 & 92.01 & 77.43  \\ \hline
        \name(4,20,5)-FT & \name+in1k & 14.19 & 86.06 & 97.46 & 91.67 & 99.60 & 93.31 & 89.17 & 75.74 & 92.89 & 83.28  \\ \hline
        \name(4,30,6)-FT & \name+in1k & 24.89 & 87.44 & 97.60 & 91.67 & 99.34 & 93.11 & 89.08 & 76.40 & 92.64 & 85.22  \\ \hline
        \name(4,50,10)-FT & \name+in1k & 40.41 & 87.55 & 98.26 & 94.78 & 99.25 & 92.77 & 89.58 & 77.23 & 93.78 & 87.11 \\ \bottomrule
    \end{tabular}
    \caption{Pre-trained transfer performance on 9 image classification tasks.}
    \label{tab:pretrained_table}
\end{table}

\renewcommand{\arraystretch}{1.1}
\setlength{\tabcolsep}{2pt}
\begin{table}[!ht]
\scriptsize
    \centering
    \begin{tabular}{l|l|l|l|l|l|l|l|l|l|l|l}
    \hline
        Network & Init. & \#\typo{Param(M)} & CIFAR100 & CIFAR10 & Caltech101 & Flowers & Cars & Aircraft & DTD & Pets & CUB  \\ \hline
        ResNet-18 & Rand & 11.69 & 77.08 & 94.03 & 64.66 & 90.94 & 90.11 & 83.71 & 61.93 & 79.10 & 75.85  \\ \hline
        ResNet-50 & Rand & 23.51 & 78.63 & 94.17 & 65.61 & 87.88 & 88.51 & 79.65 & 62.33 & 78.09 & 75.17  \\ \hline
        ResNet-101 & Rand & 42.52 & 79.87 & 94.81 & 73.33 & 87.84 & 88.21 & 78.67 & 62.63 & 76.22 & 75.35  \\ \hline
         ResNeXt 50 32x4d & Rand & 25.03 & 78.34 & 95.13 & 72.48 & 90.51 & 89.64 & 81.06 & 63.11 & 75.40 &  75.77  \\ \hline
        ResNeXt 101 32x8d Large & Rand & 88.79 & 80.23 & 96.06 & 74.61 & 91.77 & 90.88 & 88.23 & 65.51 & 77.82 & 77.61  \\ \hline
         MobileNet v3 Large 0.75 & Rand & 4.00 & 75.26 & 93.68 & 67.71 & 89.93 & 88.58 & 81.33 & 63.66 & 78.50 & 75.08  \\ \hline
        MobileNet v3 Large 1.0 & Rand & 5.40 & 75.50 & 94.07 & 73.11 & 91.88 & 89.02 & 82.69 & 63.14 & 80.12 & 75.44  \\ \hline
        RegNetY-800m & Rand & 6.30 & 79.53 & 95.35 & 73.55 & 91.22 & 90.08 & 82.14 & 62.88 & 78.61 & 76.53  \\ \hline
        RegNetY-1.6GF & Rand & 9.19 & 80.07 & 95.53 & 73.82 & 91.89 & 89.64 & 82.82 & 61.31 & 79.86 & 76.37  \\ \hline
        RegNetY-3.2GF & Rand & 20.60 & 80.81 & 96.05 & 72.38 & 90.82 & 89.33 & 78.44 & 60.70 & 80.28 & 74.94  \\ \hline
        RegNetY-8GF & Rand   & 39.20  & 81.34 & 94.61 & 74.88 & 88.73 & 88.57 & 78.21 & 61.14 & 77.08 & 75.03  \\ \hline
        RegNetY-16GF & Rand   & 83.59  & 82.35 & 95.61 & 77.23 & 92.73 & 90.26 & 81.21 & 63.67 & 78.08 & 79.03  \\ \hline
        RegNetY-32GF & Rand   & 145.05  & 82.35 & 95.61 & 77.23 & 92.73 & 90.26 & 81.21 & 63.67 & 78.08 & 79.03  \\ \hline
       Swin-T & Rand & 28.29 & 61.12 & 53.55 & 46.65 & 5.32 & 60.73 & 68.40 & 12.13 & 52.69 & 42.44  \\ \hline
        Swin-S & Rand & 49.61 & 67.83 & 55.61 & 47.14 & 13.51 & 51.64 & 67.69 & 25.08 & 43.17 & 49.53  \\ \hline
         Swin-B & Rand & 57.77 & 70.96 & 56.05 & 55.43 & 13.91 & 62.70 & 68.79 & 36.77 & 51.21 & 80.48  \\ \hline
         Swin-L &Rand & 196.53 & 72.91 & 64.28 & 56,32 & 24.26 & 59.92 & 64.91 & 59.58 & 66.91 & 76.47  \\ \hline
           ViT-T              & Rand            & 5.72          & 52.94         & 79.53 & 39.55 &   84.35  & 14.01 &16.82 &  46.65 & 32.69 & 28.36  \\ \hline
        ViT-S              & Rand            &  22.88       &   58.36      &    84.16 &   43.28   &    87.28   & 19.52 & 18.18 &  50.25 &  37.26 &    33.35  \\ \hline
        ViT-B              & Rand            &  86.86 & 63.26 & 85.63          &    45.77 & 88.27       &21.36 &  19.98 & 51.37 & 38.46 & 36.17     \\ \hline
         ViT-L & Rand & 304.72  & 67.26 & 87.71  & 47.39&  88.12 & 30.18 & 23.57 & 52.38 & 40.07 & 38.05 \\\hline
         \hline
        \name(4,10,3)-FT & \name & 7.64 & 82.67 & 95.84 & 75.02 & 97.72 & 89.88 & 82.09 & 63.82 & 80.18 & 70.93  \\ \hline
        \name(4,20,5)-FT & \name & 14.19 & 83.10 & 96.32 & 80.19 & 98.44 & 92.86 & 87.34 & 65.52 & 83.07 & 74.12  \\ \hline
        \name(4,30,6)-FT & \name & 24.89 & 84.05 & 96.42 & 82.19 & 98.61 & 93.35 & 88.86 & 70.84 & 86.9 & 79.81  \\ \hline
        \name(4,50,10)-FT & \name & 40.41 & 84.25 & 97.07 & 84.66 & 99.5 & 93.97 & 88.92 & 71.85 & 88.69 & 80.77    \\ \bottomrule
    \end{tabular}
    \caption{Randomly-initialized transfer performance on 9 image classification tasks.}
    \label{tab:rand_table}
\end{table}
\typon{\section{Extension to Other Vision Tasks}}

\typon{Apart from classification, \name~can be indeed applied to other vision tasks. There are several advantages of DeRy when it is applied to downstream tasks.}

\typon{\textbf{First}, as \name~directly searches for a general backbone, we may readily apply the same network to other tasks without any hassle.
\textbf{Second}, the training-free proxy of NASWOT does not depend on the ground-truth label and is therefore label-agnostic. It enables us to assemble new networks on any task and any modality of input.
\textbf{Third}, \name~is highly computationally efficient; it only requires several hours to search for the optimal structure on a large-sized dataset.} 

\typon{We look forward to extending DeRy to other vision tasks in our future work.}

\section{Implementation details}

\subsection{Source Code}
We implement the model training with the mmClassification\footnote{https://github.com/open-mmlab/mmclassification} framework alongside  the Pytorch backend. To support the extraction of sub-model, we require the Pytorch version $>1.10$ with the support for \texttt{torch.fx}. The NASWOT score is taken from the official implementation\footnote{https://github.com/BayesWatch/nas-without-training}. We apply \texttt{distributed training} and \texttt{FP16 half-precision training} to accelerate the model training and save the GPU memory. The code is documented and attached in a zipped file.

\subsection{Stitching Layer Structure}
Our stitching layer structures are shown in Table~\ref{tab:Stitching Layer}. For any two network blocks with different input-output sizes, we use one layer of \texttt{Norm}-\texttt{Conv}$1\times1$-\texttt{Activation} to align the dimensions and feed the output of the previous block to the next.

\begin{table}[H]
    \centering
    \begin{tabular}{l|l}
    \toprule
        Mode & Structure  \\
         \midrule
        CNN-to-CNN & \texttt{BN}    \rightarrow\texttt{Conv}$1\times1$ \rightarrow\texttt{LeakyReLU} \\
        CNN-to-Trans/MLP & \texttt{BN} \rightarrow \texttt{Conv}$1\times1$ \rightarrow \texttt{LeakyReLU} \rightarrow \texttt{Flatten}\\
        Trans/MLP-to-CNN & \texttt{Reshape(H,W)} \rightarrow \texttt{BN} \rightarrow \texttt{Conv}$1\times1$ \rightarrow \texttt{LeakyReLU} \\
        Trans/MLP-to-Trans/MLP & \texttt{LN} \rightarrow \texttt{Linear} \rightarrow \texttt{LeakyReLU} \\
         \bottomrule
    \end{tabular}
    \caption{Stitching Layer Structure.}
    \label{tab:Stitching Layer}
\end{table}

\subsection{Node Definition}
Since \name~assumes that all models are line graphs, we need to manually specify the axiom node in each neural network. \typon{In fact, not every operation in the DNN can be treated as an atomic node. Consider a \texttt{Conv->ReLU} with skip-connection; we cannot make the single convolution layer as our node because a skip-connection breaks the line graph assumption. \name, in this current form, can not cut off multiple parallel paths at the same time. Therefore, we need to specify the node in each network. For example, a ViT-B contains 12 transformer blocks and hence has 12 nodes, and a ResNet-18 has 8 residual blocks and hence has 8 nodes.}. The node definition can be found in our source code at file
\begin{center}
    \texttt{blocklize/block\_meta.py}
\end{center}
 
\subsection{Dataset and Pre-processing}
\textbf{Dataset.} We evaluated the \name~models on 10 image classification datasets in Table~\ref{tab:dataset}. These datasets cover a wide range of image classification tasks, including 4 object classification tasks CIFAR-10~\cite{krizhevsky2009learning}, CIFAR-100~\cite{krizhevsky2009learning}, Caltech-101~\cite{fei2004learning} and ImageNet1k~\cite{russakovsky2015imagenet}; 5 fine-grained classification tasks Flower-102~\cite{nilsback2008automated}, Stanford Cars~\cite{KrauseStarkDengFei-Fei_3DRR2013}, FGVC Aircraft\cite{maji13fine-grained}, Oxford-IIIT Pets~\cite{parkhi12a}, and CUB-Bird~\cite{cubbird} alongside 1 testure classification task DTD~\cite{cimpoi14describing}. The evaluation metric is either the top-1 accuracy or the per-class mean accuracy, as listed in Table~\ref{tab:dataset}.

\begin{table}[H]
    \centering
    \begin{tabular}{l|l|l|l|l}
    \toprule
        Dataset & \#Classes& \#Train & \#Test & Accuracy metric \\ \hline
        CIFAR-10 & 10 & 50,000 & 10,000 &top-1 \\ \hline
        CIFAR-100 & 100 & 50,000 & 10,000 &top-1 \\ \hline
        Stanford Cars & 196 & 8,144 & 8,041 & top-1\\ \hline
        FGVC Aircraft &100& 6,667 & 3,333 & mean per-class\\ \hline
        Describable Textures (DTD) &47 & 3,760 & 1,880  & top-1\\ \hline
        Oxford-IIIT Pets & 37 & 3,680 & 3,669 & mean per-class \\ \hline
        Caltech-101 & 102 & 3,060 & 6,084  & mean per-class\\ \hline
        Oxford 102 Flowers &102& 2,040 & 6,149 & mean per-class \\ \hline
        CUB-200-Bird & 200& 5,994 & 5,794 & top-1 \\ \hline\hline
        ImageNet1k & 1000& 1,281,167 & 50,000  & top-1 \\ 
        \bottomrule
    \end{tabular}
    \caption{Statistics and evaluation metric of datasets.}
    \label{tab:dataset}
\end{table}

\textbf{Data Pre-processing.} Following previous works~\cite{kolesnikov2020big, dosovitskiy2020image}, we train and evaluate on all datasets at the image resolution of $224 \times 224$ to align the pre-training and fine-tuning input size. For CIFAR-10 and CIFAR-100 images with original size of $32 \times 32$, we first resize them to $224\times 224$. For the other 8 datasets, a crop of random resize ratio $r \in [0.08, 1.0]$ of the original image size and a random aspect ratio $a \in [0.75, 1.33]$ of the original aspect ratio are applied to each training image, and we resize the crop to the $224\times 224$. We apply the same data augmentation on 9  transfer learning evaluations, with a random horizontal image flip with probability of 0.5, a RandAug~\cite{cubuk2020randaugment}, a random Gaussian noise with maximum kernel size of 10 and probability of 0.1, a RandomErasing~\cite{zhong2020random} with a relative area size $s \in [0.02,0.33]$ and probability of 0.2, a Mixup~\cite{zhang2017mixup} operation and a CutMix~\cite{yun2019cutmix} operation. On the ImageNet evaluation, we utilize a similar data augmentation, just without the Gaussian noise and RandomErasing. 


\subsection{Hyper-parameter Setting}
\paragraph{ImageNet.} Following recent training recipe for ImageNet, we train all models using the following setting
\begin{itemize}
    \item \textbf{Batch Size.} We utilize 8 GPUs for distributed training, each contains 128 samples. The overall \texttt{batch-size} is 1024.
    \item \textbf{Optimizer.} The networks are  optimized with AdamW~\cite{loshchilov2017decoupled} with initial learning rate 0.001, weight decay of 0.05. Weight decay is not applied to normalization layers and bias term.
    \item \textbf{Training Time.} The models are optimized for 100 epochs for \textsc{Short-training} and 300 epochs for \textsc{Full-training}. 
    \item \textbf{Learning rate schedule.} We apply the cosine learning decay, with warm-up of 5 epochs. 
    \item \textbf{Exponential Moving Average (EMA).}  We utilize EMA technique with decay 0.9995.
    \item \textbf{Label Smoothing.}  We apply label smoothing with $\epsilon=0.1$.
\end{itemize}

\paragraph{Nine Transfer Learning Tasks.} We apply grid search to fine the best hyper-parameter setting for each model-task combination. 
\begin{itemize}
    \item \textbf{Batch Size.} The \texttt{batch-size} is selected from $\{64, 128, 256\}$.
    \item \textbf{Optimizer.} The networks are either optimized with AdamW~\cite{loshchilov2017decoupled} with initial learning rate of $\frac{0.0005}{512} \times$ \texttt{batch-size} or SGD with initial learning rate of \{$0.1, 0.01$\} and Nesterov momentum of 0.9.
    \item \textbf{Training Time.} The models are optimized for \{$20k, 40k$\} iterations. 
    \item \textbf{Learning rate schedule.} We apply the cosine learning decay, without warm-up. 
    \item \textbf{Exponential Moving Average (EMA).}  We utilize EMA technique with decay 0.9995.
    \item \textbf{Label Smoothing.}  We apply label smoothing with $\epsilon=0.1$.
\end{itemize}

{\small
\bibliographystyle{plain}
\bibliography{egbib}}